\documentclass[manuscript]{acmart}
\usepackage{algorithm}
\usepackage{algpseudocode}
\AtBeginDocument{%
  }

\usepackage{amsmath,amsfonts}
\usepackage{xcolor}
\usepackage{siunitx}
\usepackage{moreverb}
\usepackage{mathtools}
\usepackage{comment}
\usepackage{graphicx}
\usepackage{tikz}
\usepackage{subfigure}
\usepackage{framed}
\usepackage{matlab-prettifier}
\usepackage{booktabs}
\usepackage{array}
\usepackage{colortbl}
\usepackage{hyperref}
\usepackage{listings}
\usepackage{enumitem}
\usepackage{threeparttable}
\usepackage{multirow}
\usepackage{fancyvrb}
\usepackage{varwidth}

\usepackage{color} %red, green, blue, yellow, cyan, magenta, black, white
\definecolor{mygreen}{RGB}{28,172,0} % color values Red, Green, Blue
\definecolor{mylilas}{RGB}{170,55,241}
\definecolor{lightblue}{rgb}{0.9, 0.95, 1.0} % Light blue for Target format columns
\definecolor{lightgreen}{rgb}{0.9, 1.0, 0.9} % Light green for Rounding modes columns

\lstdefinestyle{PythonCode}{
    language=Python,
    basicstyle=\ttfamily\tiny, % Use a typewriter font, slightly smaller
    keywordstyle=\color{teal}\bfseries, % Keywords in bold teal (e.g., class, def, return)
    morekeywords=[2]{__init__, forward, super}, % Custom keywords for functions in purple
    keywordstyle=[2]=\color{purple}\bfseries, % Style for the second keyword group (functions)
    stringstyle=\color{orange}, % Strings in orange
    commentstyle=\color{gray}\itshape, % Comments in italic gray
    identifierstyle=\color{blue}, % Variables/identifiers in blue
    numbers=left, % Line numbers on the left
    numberstyle=\tiny\color{gray}, % Line numbers are tiny and gray
    stepnumber=1, % Number every line
    numbersep=5pt, % Space between numbers and code
    showspaces=false,
    showstringspaces=false,
    showtabs=false,
    tabsize=4, % Tab size
    breaklines=true, % Break long lines
    breakatwhitespace=true, % Break only at whitespace
    backgroundcolor=\color{lightgray!10}, % Very light gray background for subtle effect
    captionpos=b, % Caption below the code
}

\begin{document}

\title{\texttt{pychop}: Emulating Low-Precision Arithmetic in Numerical Methods and Neural Networks}

\author{Erin Carson}
\thanks{The first author was funded by the Charles University Research Centre program (No. UNCE/\-24/\-SCI/005). The first and second authors were funded by the European Union (ERC, inEXASCALE, 101075632). Views and opinions expressed are those of the authors only and do not necessarily reflect those of the European Union or the European Research Council. Neither the European Union nor the granting authority can be held responsible for them.}
\email{carson@karlin.mff.cuni.cz}
\orcid{0000-0001-9469-7467}
\affiliation{%
  \institution{Department of Numerical Mathematics, Charles University}
 \city{Prague}
  \country{Czech Republic}
}

\author{Xinye Chen}
\orcid{0000-0003-1778-393X}
\email{xinye.chen@lip6.fr}
\affiliation{%
  \institution{LIP6, Sorbonne Université, CNRS}
 \city{Paris}
  \country{France}
}

\renewcommand{\shortauthors}{Carson and Chen}

\begin{abstract}
Motivated by the growing demand for reduced-precision arithmetic in computational science, we exploit lower-precision emulation in Python{--}widely regarded as the dominant programming language for numerical analysis and machine learning. Low-precision paradigms have revolutionized deep learning by enabling more efficient computation and reduced memory footprint while maintaining model fidelity. To better enable numerical experimentation with and exploration of reduced-precision computation, we developed \texttt{pychop}, which supports customizable floating-point formats and a comprehensive set of rounding modes in Python, allowing users to benefit from fast, reduced-precision emulation in numerous applications. \texttt{pychop} also provides flexible interfaces for array and tensor backends, enabling efficient reduced-precision emulation on both CPUs and GPUs for neural network deployment.

In this paper, we offer a comprehensive exposition of the design and applications of \texttt{pychop}. Furthermore, we present empirical results on reduced-precision emulation for image classification and object detection using published datasets, illustrating the sensitivity to low precision and delivering valuable insights into its quantization-aware training and post-quantization impacts. Establishing itself as a foundational tool for advancing mixed-precision algorithms,  \texttt{pychop} enables in-depth investigations into the effects of numerical precision in scientific computing and deep learning deployment,  facilitating the development of novel hardware accelerators.
\end{abstract}





\maketitle

\section{Introduction}
\label{sec:introduction}
The increasing support of reduced-precision arithmetic in hardware architectures has triggered a resurgence of interest in mixed-precision algorithms, particularly within the fields of numerical analysis and deep learning. There are numerous benefits to using numerical formats with lower precision than single precision primarily because they require less memory bandwidth. The mixed-precision algorithms, which strategically combine reduced-precision and high-precision computations, have emerged as a heating topic of research due to their capability to enhance algorithmic performance across various domains{--}most notably energy efficiency, data transfer, and arithmetic speedup{--}while maintaining acceptable levels of numerical accuracy and stability \citep{higham2022mixed}. This paradigm exploits the inherent trade-offs between computational cost and precision, addressing the escalating demands of large-scale numerical simulations and machine learning applications. The ability to tailor precision to specific computational tasks reduces energy consumption, memory footprint, and processing times, making mixed-precision computing an indispensable role in the era of exascale computing.

It is known that the increasing size of neural networks typically improves model generalization ability and accuracy at the cost of memory and compute resources for model deployment. The advance of mixed-precision training has been a milestone for deep learning efficiency. \citet{micikevicius2018mixed} demonstrated that fp16 (half-precision, 16 bits total with 5 exponent and 10 significand bits) could handle weights and activations while preserving gradient precision with fp32 accumulations, yielding a speedup of $2-3 \times$ on NVIDIA Volta GPUs. This approach has been integrated into PyTorch \citep{NEURIPS2019_9015} and implemented via Automatic Mixed Precision (AMP)\footnote{https://pytorch.org/docs/stable/amp.html}, which automates training in low precision with gradient scaling, lowering memory consumption by approximately 50\% for models like ResNet-50 \citep{7780459}. However, AMP’s dependence on hardware-supported fp16 hinders its flexibility to other floating-point formats, a constraint our emulator eliminates by supporting arbitrary precision configurations. Further, \cite{perez2023training} propose the training method via the scalings for fp8 (8-bit floating-point with two formats \citep{noune20228, micikevicius2022}{--}5 exponent bits and 2 significand bits for e5m2 and 4 exponent bits and 3 significand bits for e4m3) linear layers by dynamically updating per-tensor scales for the weights, gradients and activations, validating an acceptable performance of GPT and Llama 2 deployed with fp8 precision across model sizes ranging from 111M to 70B.

Central to this research domain is the advance of integer quantization methodologies,  including binary, ternary, and 4- to 8-bit schemes \citep{xi2023training, 10.5555/3692070.3694288}, which significantly compress model parameters and intermediate representations while maintaining performance \citep{jacob2018quantization}. \cite{xi2023training} introduces a training method for transformers with all matrix multiplications
implemented with 4-bit integer arithmetic. These methods are crucial for minimizing memory usage and computational demands, rendering them especially valuable for deploying DNNs on edge devices and other hardware with limited resources. To make the compressed information useful, it is critical to devise quantization strategies that explicitly maintain similarity between the original and quantized representations, a challenge that necessitates sophisticated similarity-preserving algorithms \citep{jacob2018quantization,10509805} that enables more efficient training and deployment of larger neural networks.

\begin{table}[ht]
    \caption{Key parameters of seven floating-point formats; $u$ denotes the unit roundoff corresponding to the precision, $x_{\min}$ denotes the smallest positive normalized floating-point number, $x_{\max}$ denotes the largest floating-point number, $t$ denotes the number of binary digits in the significand (including the implicit leading bit), $e_{\min}$ denotes the exponent of $x_{\min}$, and $e_{\max}$ denotes the exponent of $x_{\max}$. The last two columns give the number of exponent bits and significand bits (excluding the implicit bit).}
    \label{table:unitroundoff1}
    \centering
    \small
    \setlength{\tabcolsep}{4pt}
    \begin{tabular}{l c c c r r r r r}
        \toprule
        & $u$ & $x_{\min}$ & $x_{\max}$ & $t$ & $e_{\min}$ & $e_{\max}$ & exp.\ bits & sig.\ bits \\
        \midrule
        NVIDIA quarter precision (e4m3)
            & $6.25 \times 10^{-2}$
            & $1.5625 \times 10^{-2}$
            & $2.4 \times 10^{2}$
            & 4 & -6 & 7 & 4 & 3 \\
        NVIDIA quarter precision (e5m2)
            & $1.25 \times 10^{-1}$
            & $6.10 \times 10^{-5}$
            & $5.73 \times 10^{4}$
            & 3 & -14 & 15 & 5 & 2 \\
        bfloat16 (bf16)
            & $3.91 \times 10^{-3}$
            & $1.18 \times 10^{-38}$
            & $3.39 \times 10^{38}$
            & 8 & -126 & 127 & 8 & 7 \\
        half precision (fp16)
            & $4.88 \times 10^{-4}$
            & $6.10 \times 10^{-5}$
            & $6.55 \times 10^{4}$
            & 11 & -14 & 15 & 5 & 10 \\
        TensorFloat-32 (tf32)
            & $9.77 \times 10^{-4}$
            & $1.18 \times 10^{-38}$
            & $3.40 \times 10^{38}$
            & 11 & -126 & 127 & 8 & 10 \\
        single (fp32)
            & $5.96 \times 10^{-8}$
            & $1.18 \times 10^{-38}$
            & $3.40 \times 10^{38}$
            & 24 & -126 & 127 & 8 & 23 \\
        double (fp64)
            & $1.11 \times 10^{-16}$
            & $2.23 \times 10^{-308}$
            & $1.80 \times 10^{308}$
            & 53 & -1022 & 1023 & 11 & 52 \\
        \bottomrule
    \end{tabular}
\end{table}

We design a reduced-precision emulation software called \texttt{pychop} for Python, offering a highly customizable setting for users to simulate arbitrary reduced-precision arithmetic{--}floating-point, fixed-point, and integer{--}with a diverse of rounding modes. This emulator transcends the constraints of fixed hardware {by allowing users to define custom precision formats{--}specifying exponent and significand bits{--}and to select from rounding modes, e.g., round to nearest, round up / down,  round toward zero,  round toward odd, and two stochastic variants (proportional and uniform probability).} The software promotes comprehensive precision format support, flexibility, and a beginner-friendly interface:
\begin{itemize}
 \item \textbf{Unmatched Flexibility}: Emulating both standard (see \tablename~\ref{table:unitroundoff1}) and customized numerical formats for reduced-precision emulation, enabling researchers to prototype hypothetical hardware or explore theoretical precision boundaries without physical constraints.
 \item \textbf{Precision Granularity}: Providing precise control over numerical representation and supporting soft error emulation in arithmetic operations, which is critical for dissecting the impact of quantization on both numerical algorithms and deep learning methods, e.g., gradient updates, weight distributions, and activation ranges in neural networks.
 \item \textbf{Seamless Integration}: Offering direct emulators into PyTorch layers, enabling practitioners to experiment with mixed-precision training and inference pipelines with minimal overhead. The user-friendly API enables users to deploy quantization-aware and post-quantization strategies easily.
 \item \textbf{Rounding Mode Exploration}: Supporting a diverse collection of rounding modes is available, enabling empirical evaluation of their roles in numerical stability.  Our stochastic rounding is high-performance implementation, which shows empirically comparable performance to deterministic rounding.
 % \item  \textbf{}
\end{itemize}

Using \texttt{pychop}, we present a comprehensive evaluation of training neural networks with quantization-aware training and post-quantization strategies in image classification tasks and object detection tasks, which offers insights into how different precision with rounding modes behave with respect to performance gains. Our software and experimental code are publicly available at \url{https://github.com/inEXASCALE/pychop}.

This paper is organized as follows: Section~\ref{sec:related_work} reviews prior work on precision emulation software \texttt{pychop}, identifying gaps that our solution addresses; Section~\ref{sec:methodology} describes the implementation and usage in detail; and Section~\ref{sec:exps} presents simulated experiments that collectively demonstrate the emulator's performance in MATLAB and Python and its value in advancing mixed-precision emulation for deep learning applications, namely image classification and object detection. Section~\ref{sec:conclusion} concludes the paper and outlines future work.

\section{Related Work}
\label{sec:related_work}

\begin{table}[ht]
    \centering
    \setlength\tabcolsep{2.0pt}
    \footnotesize
    \caption{Software for simulating reduced-precision arithmetic. The table abbreviates long built-in format lists; additional OCP MX, BFP, and Flexpoint support is described in the note below.}
    \label{tab:software-packages}
    \begin{tabular}{l c c >{\columncolor{lightblue}}c >{\columncolor{lightblue}}c >{\columncolor{lightblue}}c >{\columncolor{lightblue}}c *{7}{>{\columncolor{lightgreen}}c} c c c c}
        \toprule
        \multirow{2}{*}{\textbf{Package}} & \multirow{2}{*}{\textbf{Lang.}} & \multirow{2}{*}{\textbf{Storage}} & \multicolumn{4}{c}{\textbf{Target format}} & \multicolumn{7}{c}{\textbf{Rounding modes}} & \multirow{2}{*}{\textbf{FPQ}} & \multirow{2}{*}{\textbf{IQ}}& \multirow{2}{*}{\textbf{NN}} & \multirow{2}{*}{\textbf{STE}}\\
        \cmidrule(lr){4-7} \cmidrule(lr){8-14}
        & & & $\mathbf{p}$ & $\mathbf{e}$ & $\mathbf{s}$ & \textbf{Built-in} & \rotatebox{90}{RNE} & \rotatebox{90}{RNZ} & \rotatebox{90}{RNA} & \rotatebox{90}{RZ} & \rotatebox{90}{RUD} & \rotatebox{90}{RO} & \rotatebox{90}{SR} & \\
        \midrule
        \texttt{GNU MPFR} \cite{10.1145/1236463.1236468} & C & \text{custom} & A & A & O & & $\checkmark$ & & $\checkmark$ & $\checkmark$ & $\checkmark$ & & \\
        \texttt{SIPE} \cite{lefevre:hal-00864580} & C & \text{multiple} & R & S & Y & & $\checkmark$ & & & $\checkmark$ & & & \\
        \texttt{rpe} \cite{gmd-10-2221-2017} & Fortran & \text{fp64} & R & B & B & \text{fp16} & $\checkmark$ & & & & & & \\
        \texttt{FloatX} \cite{10.1145/3368086} & C++ & \text{fp32/fp64} & R & S & Y & & $\checkmark$ & & & & & & \\
        \texttt{FlexFloat} \cite{8556098} & C++ & \text{fp32/fp64} & R & S & Y & & $\checkmark$ & & & & & & \\
        \texttt{INTLAB} \cite{Rump1999} & MATLAB & \text{fp64} & R & S & Y & & $\checkmark$ & & & $\checkmark$ & $\checkmark$ & & \\
        \texttt{Chop} \cite{doi:10.1137/19M1251308} & MATLAB & \text{fp32/fp64} & R & S & F & \text{fp16/bf16} & $\checkmark$ & & & $\checkmark$ & $\checkmark$ & & $\checkmark$ \\
        \texttt{CPFloat} \cite{10.1145/3585515} & C & \text{fp32/fp64} & R & S & F & \text{fp16/bf16/tf32} & $\checkmark$ & $\checkmark$ & $\checkmark$ & $\checkmark$ & $\checkmark$ & $\checkmark$ & $\checkmark$ \\
        \texttt{QPyTorch} \cite{9463516} & Python & \text{fp32} & R & S & N & & $\checkmark$ & $\checkmark$ & & & & & $\checkmark$  & && $\checkmark$& $\checkmark$ \\
        \texttt{gfloat} \cite{gfloat2024} & Python & \text{custom} & A & A & F & \text{fp16/bf16/fp8} & $\checkmark$ & & $\checkmark$ & $\checkmark$ & $\checkmark$ & & $\checkmark$ & & & & \\
        \texttt{pychop} & Python / MATLAB & \text{fp32/fp64} & R & S & F & \text{fp16/bf16/tf32/fp8} & $\checkmark$ & $\checkmark$ & $\checkmark$ & $\checkmark$ & $\checkmark$ & $\checkmark$ & $\checkmark$ & $\checkmark$& $\checkmark$ & $\checkmark$& $\checkmark$ \\
        \bottomrule
    \end{tabular}
    \begin{flushleft}
        \small
        The columns are categorized as follows: (i) \textbf{Package, Lang., Storage}: The name of the package, its primary programming language, and the supported storage formats. (ii) \textbf{Target format parameters}: $p$: Number of bits of precision in the significand{--}arbitrary (A) or restricted to the storage format's significand (R); $e$: Exponent range{--}arbitrary (A), a sub-range of the storage format (S), or a sub-range only for built-in types (B); $s$: Support for subnormal numbers{--}supported (Y), not supported (N), supported only for built-in types (B), supported by default but can be disabled (F), or not supported by default but can be enabled (O); \textbf{Built-in}: floating-point formats natively built into the system.
        \texttt{gfloat} additionally supports OCP/MX formats.  Current \texttt{pychop} releases, including version 0.6.0, support OCP MX, BFP, and Flexpoint-style formats; 
        (iii) \textbf{Rounding modes}: Supported modes include round-to-nearest with ties-to-even (RNE), ties-to-zero (RNZ), ties-to-away (RNA), round-toward-zero (RZ), round up/down (RUD), round-to-odd (RO), and stochastic rounding variants (SR). A $\checkmark$ indicates full support. (iv) \textbf{FPQ}: support fixed-point quantization. (v) \textbf{IQ}: support integer quantization. (vi) \textbf{NN}: support neural network quantization. (vii) \textbf{STE}: support Straight-Through Estimator, which permits gradients to propagate through during the backward pass that enables the continuation of the backpropagation algorithm.
    \end{flushleft}
\end{table}

% floating-point
Low-precision arithmetic has emerged as a pivotal technique for optimizing computational efficiency in scientific computing and machine learning applications, where reduced precision can significantly lower resource demands while maintaining acceptable accuracy \citep{doi:10.1137/19M1251308}. Several software libraries have been developed to emulate reduced-precision arithmetic, each with distinct capabilities tailored to specific use cases. In this section, we discuss the software packages listed in Table~\ref{tab:software-packages}\footnote{The table's design follows \cite{10.1145/3585515}}, highlighting their strengths and limitations in the context of reduced-precision arithmetic simulation.

GNU MPFR \citep{10.1145/1236463.1236468} is a C library for simulating multiple-precision floating-point computations with guaranteed correct rounding. It excels in scenarios requiring arbitrary precision for both the significand ($p$) and exponent range ($e$). It is a preferred choice for numerical simulations demanding high accuracy, such as symbolic computation and numerical analysis. However, its default lack of subnormal number support ($s$, denoted as O) requires explicit user configuration, which can complicate workflows. Besides, GNU MPFR does not offer built-in floating-point formats and is not suited for neural network training, limiting its capability in reduced-precision emulation in deep learning.

\texttt{SIPE} \citep{lefevre:hal-00864580}, another C-based library, focuses on very reduced-precision computations with correct rounding. It supports multiple storage formats, restricted significand ($p$), and a restricted exponent range ($e$), while fully supporting subnormal numbers ($s$). SIPE implements RNE and RZ rounding modes for reduced-precision simulations which is suitable for exploring numerical stability and precision trade-offs in low-bitwidth computations, such as those encountered in embedded systems. Its primary limitation lies in its restricted rounding mode support and lack of built-in formats, which reduces its versatility.

\texttt{rpe} \citep{gmd-10-2221-2017}, implemented in Fortran, is tailored for emulating reduced floating-point precision in large-scale numerical simulations, such as climate modeling. It operates with restricted significand ($p$) and exponent range ($e$) limited to built-in types (B) and supports subnormal numbers only for built-in types (B). A key advantage is its native support for the fp16 format, aligning with reduced-precision hardware standards like IEEE 754 binary16. However, rpe's exclusive support for RNE rounding limits its flexibility in scenarios requiring diverse rounding strategies, and it does not support neural network training, focusing solely on numerical simulations.

\texttt{FloatX} \citep{10.1145/3368086} and \texttt{FlexFloat} \citep{8556098} are C++ libraries that provide frameworks for customized floating-point arithmetic in reduced-precision simulations. These libraries support restricted significand ((p)) and exponent range ((e)), fully accommodate subnormal numbers ((s)), and utilize fp32/fp64 storage formats for compatibility with standard representations. Their simplicity, limited to round-to-nearest-even (RNE) rounding, makes them accessible for educational purposes and prototyping. However, this restricted rounding support limits their adaptability, and neither library includes predefined formats or supports neural network training, confining their use to general-purpose reduced-precision arithmetic experimentation.  Both \texttt{FloatX} and \texttt{FlexFloat} strictly follow standard C++ conventions, including round-to-nearest, ties-to-even rounding, and default datatype casting. Although these conventions ensure compatibility with native floating-point operations, they limit users’ ability to explore diverse rounding strategies in numerical simulations. This constraint poses a notable challenge for researchers and practitioners needing flexible, application-specific rounding behaviors.

\texttt{INTLAB} \citep{Rump1999}, a MATLAB toolbox, leverages interval arithmetic to facilitate reduced-precision floating-point simulation. It uses fp64 storage, supports restricted significand ($p$) and exponent range ($e$), and fully supports subnormal numbers ($s$). INTLAB provides RNE, RZ, and RUD rounding modes, offering moderate flexibility for numerical computations in MATLAB environments, such as verified computing. Its lack of built-in formats and optimization for neural network training limits its scope, positioning it as a tool for reliable numerical analysis rather than machine learning.

\texttt{Chop} \citep{doi:10.1137/19M1251308}, another MATLAB library, enables reduced-precision arithmetic simulation with fp32/fp64 storage, restricted significand ($p$), and exponent range ($e$), alongside flexible subnormal number support (F). It supports built-in fp16 and bf16 formats, aligning with modern hardware standards, and implements RNE, RZ, RUD, and stochastic rounding (SR). The inclusion of SR is particularly valuable for simulating quantization effects, which are critical in machine learning research. Despite this, chop does not directly support neural network training and inference, and it is limited to the MATLAB environment, restricting its application to numerical experimentation and analysis of quantization impacts in machine learning.

CPFloat \citep{10.1145/3585515} is a C library optimized for efficient reduced-precision arithmetic simulation, supporting fp32/fp64 storage with restricted significand ($p$), and exponent range ($e$). It includes built-in fp16, bf16, and tf32 formats with flexible subnormal number support (F), enhancing compatibility with hardware-accelerated systems such as GPUs. CPFloat supports a comprehensive set of rounding modes that makes it highly versatile for observing the numerical behavior of reduced-precision arithmetic. However, CPFloat is not designed for neural network quantization but rather for general-purpose reduced-precision arithmetic in numerical algorithms.

For reduced-precision emulation in Python, \texttt{QPyTorch} \citep{9463516} is a PyTorch-based reduced-precision simulator, specifically designed for reduced-precision arithmetic for neural network training without relying on reduced-precision hardware. As mentioned in \citep{10.1145/3585515},  its principles are analogous to those of \texttt{Chop}, in that numbers are stored in binary32 before as well as after rounding, and offers RNE, RNZ, and SR rounding modes, enabling efficient training via fused CUDA kernels. However, infinities, NaNs, and subnormals are excluded for efficiency because neural network training typically doesn't involve these special values. Further, its limited rounding mode support (lacking RZ, RUD, RO) restricts quantization studies and numerical simulations. \texttt{gfloat} \citep{gfloat2024} is a Python library designed for simulating reduced-precision floating-point arithmetic. It allows experimentation with various floating-point formats in Python, including IEEE 754 and OCP/MX (Microscaling Formats) \citep{10.1145/103162.103163, Rouhani2023OCP}, and supports stochastic rounding, but does not include soft-error or fault-injection simulation features. \texttt{gfloat} supports array in NumPy, PyTorch, and JAX.

Within the scope of mixed-precision computing in deep learning, all aforementioned libraries, except \texttt{QPyTorch}, lack Straight-Through Estimator (STE) support and are thus restricted to post-training quantization (PTQ), rendering them ineffective for quantization-aware training (QAT). Our \texttt{pychop} library supports fp32/fp64 storage, built-in fp16, bf16, tf32, fp8-style formats, OCP MX microscaling formats, Block Floating Point (BFP), Flexpoint, constrained precision ($p$) and exponent ($e$), and flexible significand ($s$), with rounding modes (RNE, RNZ, RNA, RZ, RUD, RO, and SR). STE is exposed explicitly through \texttt{ChopSTE}, \texttt{ChopfSTE}, \texttt{ChopiSTE}, and quantized neural-network layers. \texttt{pychop} integrates seamlessly with NumPy, PyTorch, and  JAX{--}outperforming QPyTorch’s PyTorch-only scope{--}and offers efficient rounding with lazy optional backend imports. Its versatility and multi-framework compatibility make \texttt{pychop} a superior tool for quantized neural network training across diverse deep-learning workflows and scientific computations.

\section{The \texttt{pychop} library}
\label{sec:methodology}

Numbers in machines are often approximated with truncation using discrete representations due to the constraints arisen from finite storage. Two fundamental representations dominate this domain: \emph{floating-point representation}, which excels in representing a wide range of values with variable precision, and \emph{fixed-point representation}, which prioritizes simplicity and efficiency in constrained environments. This section formulates both representations, with a particular focus on the IEEE 754 standard for floating-point representation, and compares their theoretical and practical implications in computational tasks. Our mixed-precision emulation software is implemented as modular Python code with NumPy, PyTorch, and  JAX backends. It comprises scalar floating-point emulation through \texttt{Chop} and \texttt{FaultChop}, fixed-point quantization through \texttt{Chopf}, integer quantization through \texttt{Chopi}, and block or microscaling formats such as BFP, Flexpoint, and OCP MX. These components are integrated with \texttt{layers} interfaces that provide explicit straight-through estimators for neural-network deployment, supporting various representation formats (see \tablename~\ref{tab:rounding_modes} for details) as well as subnormal numbers. In the following, we detail its design and implementation, emphasizing its main principles and user-friendly features.

Unless otherwise stated, the API examples in this paper reflect the current \texttt{pychop} 0.6.0 interface.

\begin{table}[ht]
\centering\footnotesize
\caption{Rounding modes for the \texttt{rmode} in reduced-precision emulation.}
\label{tab:rounding_modes}
\begin{tabular}{clp{9cm}}
\toprule
\textbf{rmode} & \textbf{Rounding Mode} & \textbf{Description} \\
\midrule
1 & Round to nearest, ties to even & Rounds to the nearest representable value; in cases of equidistance, selects the value with an even least significant digit (IEEE 754 standard). \\
2 & Round toward +$\infty$ (round up) & Rounds to the smallest representable value greater than or equal to the input, directing towards positive infinity. \\
3 &  Round toward $-\infty$ (round down) & Rounds to the largest representable value less than or equal to the input, directing towards negative infinity. \\
4 & Round toward zero & Discards the fractional component, yielding the integer closest to zero without exceeding the input’s magnitude. \\
5 & Stochastic (proportional) & Employs probabilistic rounding where the probability of rounding up is proportional to the fractional component. \\
6 & Stochastic ({uniform}) & Applies probabilistic rounding with an equal probability (0.5) of rounding up or down, independent of the fractional value. \\
7 & Round to nearest, ties to zero & Rounds to the nearest representable value; in cases of equidistance, selects the value closer to zero. \\
8 & Round to nearest, ties away & Rounds to the nearest representable value; in cases of equidistance, selects the value farther from zero. \\
9 & Round toward odd & Rounds to the nearest representable odd value; in cases of equidistance, selects the odd value in the direction of the original number. \\
\bottomrule
\end{tabular}
\end{table}

\subsection{Floating-point Arithmetic}
Floating-point representation approximates real numbers using a binary format analogous to scientific notation. \texttt{pychop} emulates floating-point arithmetic by decomposing a tensor into sign, exponent, and significand components, following IEEE 754 conventions; The IEEE 754 standard, established in 1985 and revised in 2008 and 2019, provides a widely adopted framework for floating-point arithmetic, ensuring consistency across hardware and software implementations.

A floating-point number $ x $ in the IEEE 754 standard is defined based on its encoding as a tuple of three components: a sign bit, an exponent, and a mantissa (or significand). Mathematically, the value $ x $ is expressed as
\begin{equation*}
x = (-1)^s \cdot M \cdot 2^E,
\end{equation*}
where $ s \in \{0, 1\} $ is the sign bit ($ s = 0 $ for positive, $ s = 1 $ for negative), $ M $ is the mantissa, a binary fraction interpreted based on the exponent field, and $ E $ is the exponent{--}an integer derived from the stored exponent field with a bias adjustment.

The IEEE 754 standard defines several formats, with the most common being single precision and double precision. For a format with a $ k $-bit exponent and a $ p-1 $-bit mantissa fraction, the bit layout is
\begin{equation*}
[s \, | \, e \, | \, m],
\end{equation*}
where $ s $ is the 1-bit sign,  $ e $ is the $ k $-bit exponent field, and $ m $ is the $ p-1 $-bit fractional part of the mantissa{--}the total precision $ p $ includes an implicit leading bit for normalized numbers.

The interpretation of $ M $ and $ E $ depends on the value of the exponent field $ e $:
\begin{itemize}
    \item \emph{Normalized Numbers} ($ 1 \leq e \leq 2^k - 2 $): The mantissa is $ M = 1 + \sum_{i=1}^{p-1} m_i \cdot 2^{-i} $, where $ m_i $ are the bits of the fractional field, and the exponent is $ E = e - \text{bias} $, with $ \text{bias} = 2^{k-1} - 1 $. Thus
    \begin{equation*}
    x = (-1)^s \cdot \left(1 + \sum_{i=1}^{p-1} m_i \cdot 2^{-i}\right) \cdot 2^{e - \text{bias}}.
    \end{equation*}
    \item \emph{Denormalized Numbers} ($ e = 0 $): The mantissa is $ M = 0 + \sum_{i=1}^{p-1} m_i \cdot 2^{-i} $ (no implicit leading 1), and the exponent is $ E = 1 - \text{bias} $. Thus
    \begin{equation*}
    x = (-1)^s \cdot \left(\sum_{i=1}^{p-1} m_i \cdot 2^{-i}\right) \cdot 2^{1 - \text{bias}}.
    \end{equation*}
    Denormalized numbers allow representation of values closer to zero, mitigating the abrupt underflow of normalized numbers.
    \item \emph{Special Values}:
    \begin{itemize}
        \item If $ e = 2^k - 1 $ and $ m = 0 $, then $ x = (-1)^s \cdot \infty $ (infinity).
        \item If $ e = 2^k - 1 $ and $ m \neq 0 $, then $ x $ represents a Not-a-Number (NaN), used to indicate invalid operations.
        \item If $ e = 0 $ and $ m = 0 $, then $ x = (-1)^s \cdot 0 $ (signed zero).
    \end{itemize}
\end{itemize}

% \subsection{Implementation details}
Floating-point arithmetic often produces results that cannot be represented exactly within the finite precision $ p $. The IEEE 754 standard defines several rounding modes to map an exact real number $ r \in \mathbb{R}$ to a representable floating-point number $ x \in \mathbb{F}$. Let $ \lfloor r \rfloor_{\mathbb{F}} $ and $ \lceil r \rceil_{\mathbb{F}} $ denote the closest representable numbers in $ \mathbb{F} $ such that $ \lfloor r \rfloor_{\mathbb{F}} \leq r \leq \lceil r \rceil_{\mathbb{F}} $. The midpoint between two consecutive representable numbers is $ m = \frac{\lfloor r \rfloor_{\mathbb{F}} + \lceil r \rceil_{\mathbb{F}}}{2}$.

In the following, we present two fundamental modules that \texttt{pychop} offers{--}namely \texttt{Chop} and \texttt{FaultChop}{--}for rounding numbers to reduced-precision binary floating-point format. \texttt{FaultChop} features a greater set of functionalities corresponding to Nick Higham's original implementation\footnote{\url{https://github.com/higham/chop}}, with minor code optimizations for improved performance. In contrast to \texttt{Chop}, \texttt{FaultChop} retains full support for soft error simulation, whereas \texttt{Chop} simplifies certain operations to achieve greater speedup. In other words, \texttt{Chop} is designed to be a lightweight ``\texttt{FaultChop}'' with essential features and enables efficient vectorization. Generally, \texttt{Chop} is faster than \texttt{FaultChop}, which will be verified in Section~\ref{sec:exps}, but \texttt{FaultChop} includes more floating-point emulation features. As such, \texttt{Chop} is designed as a high-performance interface optimized for neural network training and large-scale arrays, while \texttt{FaultChop} is designed as a full-featured research interface supporting complete custom formats and \textit{soft-error simulation} via random bit-flipping in the significand (\texttt{flip=True}). This makes it particularly suitable for studying numerical robustness under hardware faults. We demonstrate their basic usage in Appendix~\ref{app:fpt}.

Further, different backends of \texttt{Chop} offer different features. The PyTorch and JAX backends support tensor-native execution, including GPU deployment when the underlying framework and input tensors are placed on a GPU device. The NumPy backend can leverage \texttt{Dask} \citep{swdask}, when available, to process large NumPy arrays by chunking the array with a user-defined chunk size. The chunk size determines how the array is split into smaller blocks for parallel processing, balancing memory use and scheduling overhead.

The calling of \texttt{pychop} is straightforward. In the following, we will illustrate how to use \texttt{Chop} and  \texttt{FaultChop} to emulate reduced-precision arithmetic.  The prototype of \texttt{Chop} and \texttt{FaultChop}  are separately described as below:
\begin{itemize}[leftmargin=40pt, labelsep=5pt, itemsep=10pt, topsep=20pt]

\item \begin{lstlisting}
Chop(exp_bits:int, sig_bits:int, rmode:int=1, subnormal:bool=True, chunk_size:int=800, random_state:int=42, verbose:int=0)
\end{lstlisting}
This interface facilitates precise control over the precision range and rounding behavior of floating-point operations. It enables users to specify the bitwidth for the exponent (\texttt{exp\_bits}) and significand (\texttt{sig\_bits}) of floating-point numbers. A rounding mode parameter (\texttt{rmode}), defaulting to 1, governs rounding behavior. Subnormal numbers are managed via the \texttt{subnormal} parameter, which defaults to \texttt{True}. The \texttt{chunk\_size} parameter is accepted by the unified interface and is primarily used by the NumPy/Dask implementation to choose the block size for large arrays. Additionally, a random seed (\texttt{random\_state}), defaulting to 42, can be configured to ensure reproducibility in stochastic rounding scenarios, and \texttt{verbose} controls optional diagnostic output.

\item \begin{lstlisting}
FaultChop(prec:str='h', subnormal:bool=None, rmode:int=1, flip:bool=False, explim:int=1, p:float=0.5, randfunc=None, customs:Customs=None, random_state:int=0, verbose:int=0)
\end{lstlisting}
This interface is designed to support detailed control over precision, range, and rounding behavior in floating-point operations, allowing users to specify the target arithmetic precision (\texttt{prec}, defaulting to 'h'; if \texttt{customs} is provided, the custom parameters define the format), whether subnormal numbers are supported (\texttt{subnormal}), and the rounding mode (\texttt{rmode}, with options like ``nearest'' or stochastic methods, defaulting to 1). Additional features include an option to randomly flip bits in the significand for error simulation (if \texttt{flip}, which defaults to \texttt{False}, is set to \texttt{True}, then each element of the rounded result has, with probability \texttt{p} (default 0.5), a randomly chosen bit in its significand flipped), controls for exponent limits (\texttt{explim}), and a custom random function for stochastic rounding (\texttt{randfunc}, defaulting to None). Users can also define custom precision parameters via a dataclass \texttt{Customs(emax, t, exp\_bits, sig\_bits)} where \texttt{emax} refers to the maximum value of the exponent, \texttt{t} refers to the significand bits which includes the hidden bit, and \texttt{exp\_bits} and \texttt{sig\_bits} refer to the exponent bit and significand bit which excludes the hidden bit, respectively.  \texttt{random\_state} is used to set a random seed for reproducibility, and \texttt{verbose} toggles unit-roundoff output.

\end{itemize}

We intentionally provide full support for subnormal numbers (with a user-controllable switch to enable or disable them) as a core capability, primarily to allow \texttt{pychop} to faithfully emulate the complete semantics of the IEEE 754 standard, rather than merely pursuing maximum computational speed. This is particularly important for numerical computing since subnormal support enables the gradual underflow mechanism that effectively prevents the sudden loss of precision caused by abrupt underflow, thereby significantly improving numerical stability in long-running iterations, ill-conditioned problems, or scenarios involving the accumulation of very small quantities. Surveys and relevant references can be found in \cite{10.1145/103162.103163} and \cite{higham2002accuracy}.

\begin{minipage}{0.6\textwidth}
To show the practical benefit of the support for subnormal numbers, we consider the sequential summation of a geometric series
\begin{equation*}
\sum_{k=0}^{N-1} s \cdot r^k
\end{equation*}
where the addends are chosen to straddle the normal-to-subnormal boundary.

As shown in Figure~\ref{fig:subnormal_geometric}, with parameters
$N = 1000$, $r = 0.99$ and $s=2.5 \times 10^{-6}$, enabling subnormals
allows gradual underflow prevention and disabling them (flush-to-zero)
causes early stagnation. This example highlights why faithful IEEE~754
emulation with subnormals is essential for numerical methods involving
long accumulations.\\
\end{minipage}
\hfill
\begin{minipage}{0.35\textwidth}
\centering
\includegraphics[width=\textwidth]{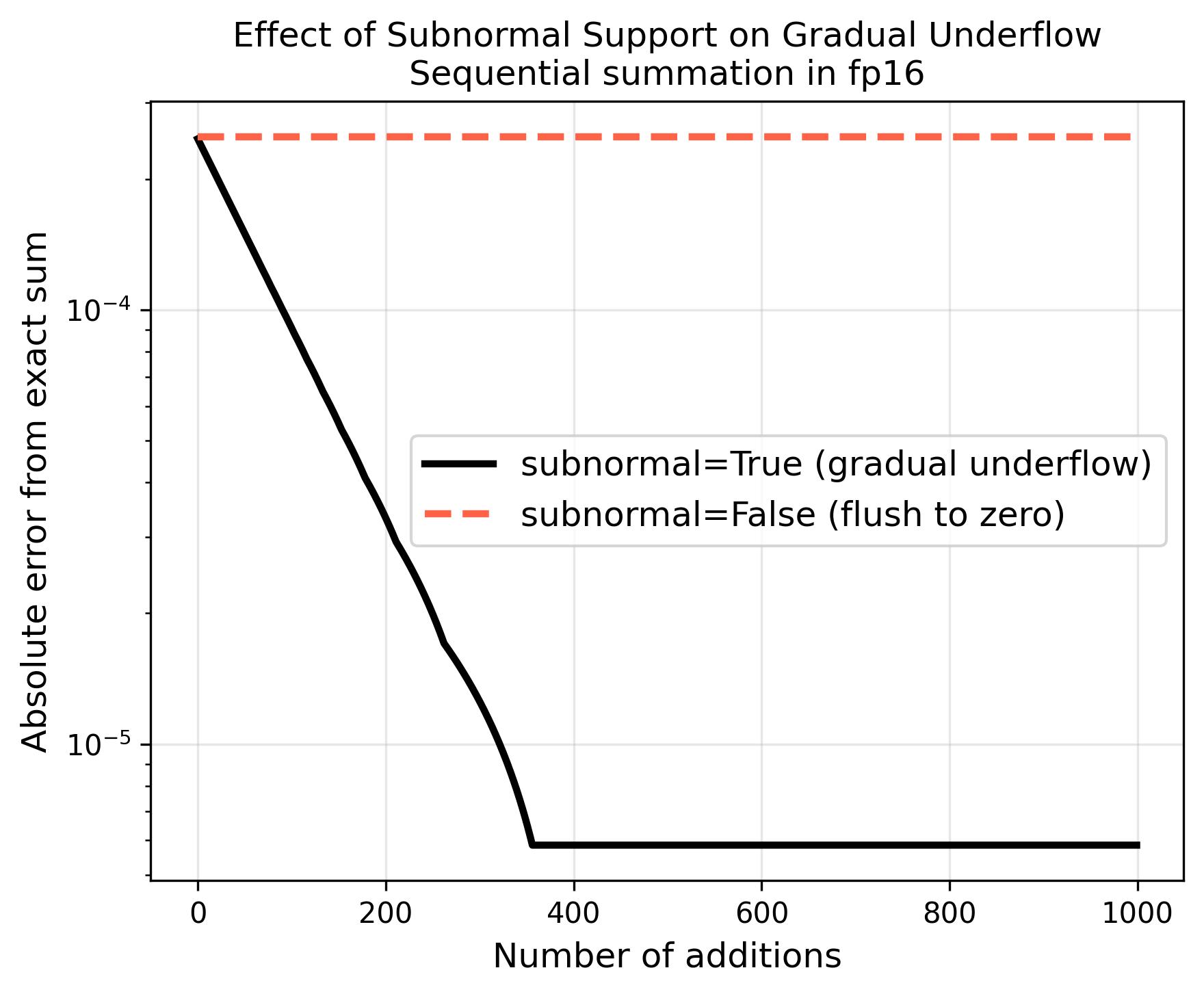}
\captionof{figure}{Effect of subnormal support in fp16.}
\label{fig:subnormal_geometric}
\end{minipage}

\subsection{Fixed-point Arithmetic}
Fixed-point is associated with numbers of a fixed number of bits, allocating a predetermined number of bits to the integer part and the fractional part, unlike floating-point representations that use an exponent and significand. The shift to fixed-point arithmetic is driven by several key advantages. First, fixed-point compute units are typically faster and significantly more efficient in terms of hardware resources and power consumption compared to floating-point units. Their smaller logic footprint allows for a higher density of compute units within a given area and power budget. Second, reduced-precision data representation reduces the memory footprint, enabling larger models to fit within available memory while also lowering bandwidth requirements. Fixed-point operations align well with digital signal processors (DSPs) and field-programmable gate arrays (FPGAs) because many lack dedicated floating-point units or optimize for fixed-point arithmetic. Collectively, these benefits enhance data-level parallelism, leading to substantial performance gains \citep{10.5555/3045118.3045303}.

Fixed-point representation provides a simpler alternative to floating-point by fixing the position of the binary point within a binary number. This conversion from floating-point numbers to fixed-point numbers employs a fixed scaling factor (implicit or explicit), enabling fractional values to be represented as integers scaled by a constant, such as $ 2^{-f}$, where $f$ is the number of fractional bits.

A fixed-point number $x$ is defined as an integer $I$ scaled by a fixed factor $2^{-f}$, where $f$ is the number of fractional bits:
\begin{equation*}
x = I \cdot 2^{-f}
\end{equation*}
where  $I$ is an integer, which may be signed (typically in two’s complement) or unsigned,  $f$ is the fixed number of fractional bits,
and $n$ is the total number of bits is, with $ n-f $ bits allocated to the integer part and $f$ bits to the fractional part.

The binary representation of $ I $ can be written as:
\begin{equation*}
I = b_{n-1} b_{n-2} \ldots b_f \cdot b_{f-1} \ldots b_0
\end{equation*}
with the binary point implicitly placed between bits $ b_f $ and $ b_{f-1} $. The numerical value is
\begin{equation*}
x = \left( \sum_{i=0}^{n-1} b_i \cdot 2^i \right) \cdot 2^{-f}.
\end{equation*}
For a signed representation using two’s complement, the most significant bit $ b_{n-1} $ is the sign bit, and the value of $ I $ is
\begin{equation*}
I = -b_{n-1} \cdot 2^{n-1} + \sum_{i=0}^{n-2} b_i \cdot 2^i.
\end{equation*}
Thus, the fixed-point value $ x $ becomes
\begin{equation*}
x = \left( -b_{n-1} \cdot 2^{n-1} + \sum_{i=0}^{n-2} b_i \cdot 2^i \right) \cdot 2^{-f}.
\end{equation*}
For an unsigned representation, the sign bit is absent, and $ I = \sum_{i=0}^{n-1} b_i \cdot 2^i $, so
\begin{equation*}
x = \left( \sum_{i=0}^{n-1} b_i \cdot 2^i \right) \cdot 2^{-f}.
\end{equation*}

% \subsubsection{Process of Fixed-point Quantization}

The process of fixed-point quantization in neural networks involves several steps. A fixed-point number is typically denoted as $ Qm.f $, where $ m $ represents the number of integer bits and $ f $ the number of fractional bits. For instance, a $ Q8.8 $ format uses 8 bits for the integer part and 8 bits for the fractional part, stored as a 16-bit integer. To convert a floating-point value $ x $ to a fixed-point value $ q $, the value is scaled and rounded according to
\begin{equation*}\label{eq:fpround}
q = \text{round}(x \cdot 2^f).
\end{equation*}

The prototype of \texttt{pychop} for fixed-point quantization is as below:
\begin{lstlisting}
Chopf(ibits:int=4, fbits:int=4, rmode:int=1)
\end{lstlisting}
where \texttt{ibits} refers to the bitwidth of integer part,  \texttt{fbits} refers to the bitwidth of fractional part, and \texttt{rmode} indicates the rounding mode used in \eqref{eq:fpround}; the supported rounding modes can be found in  \tablename~\ref{tab:rounding_modes}.

The usage example code is given in Appendix~\ref{app:fp}.

% \subsubsection{Relation to and Deep Learning Deployment}

The adoption of fixed-point quantization in deep learning offers several benefits and significant contributions that enhance its utility across various applications and plays a pivotal role in neural networks by optimizing both inference and training phases.
By leveraging fixed-point arithmetic, which uses fewer bits—such as 16-bit $ Q8.8 $ representations compared to 32-bit floats{--}this technique substantially reduces the memory footprint and accelerates multiply-accumulate (MAC) operations, which are fundamental to DNN computation. Its compatibility with hardware is a key advantage, as many edge devices like microcontrollers natively support fixed-point operations, eliminating the need for floating-point emulation and thereby boosting performance. With careful selection of the integer ($ m $) and fractional ($ f $) bit allocations, fixed-point quantization maintains accuracy close to that of floating-point models, a capability validated by research \citep{10.5555/3045390.3045690}. Ultimately, this technique has enabled the deployment of sophisticated DNNs on resource-limited platforms, significantly broadening the practical impact of AI in fields like mobile computing, real-time image processing, and autonomous navigation.

In the context of neural networks and deep learning, fixed-point quantization is applied to weights, activations, and sometimes gradients to optimize computation and storage, making it a cornerstone for efficient deployment and training.  For inference, it replaces costly floating-point operations with fixed-point arithmetic, which is natively supported by many embedded systems. In training, quantization-aware training ensures the model adapts to reduced precision, minimizing accuracy loss. Fully fixed-point training, though less common, has been studied for end-to-end optimization on fixed-point hardware. This compatibility with hardware acceleration—particularly DSPs and FPGAs, which often lack native floating-point units or optimize for fixed-point operations—bridges the gap between complex DNNs (e.g., convolutional neural networks or Transformers) and practical, low-power deployment. Fixed-point quantization thus reduces memory and computational costs while enabling the practical deployment of DNNs on edge devices, as explored in foundational works like \cite{10.5555/3045390.3045690}.

\subsection{Integer Arithmetic}
Integer quantization is a fundamental technique in digital systems to approximate real numbers using a finite set of integers, enabling efficient storage and computation in applications such as machine learning, digital signal processing, and embedded systems. Two primary approaches to quantization exist: \emph{symmetric quantization}, which balances positive and negative ranges around zero, and \emph{asymmetric quantization}, which allows unequal ranges for positive and negative values.
In principle, integer quantization maps a real number $r \in \mathbb{R} $ to a discrete integer value $x \in \mathbb{Z} $ through scaling and rounding. The process can be adapted for either symmetric or asymmetric quantization depending on the range and scaling strategy. Consider a real number $r$ within a specified range $[r_{\text{min}}, r_{\text{max}}]$. The goal is to represent $r$ using $n$-bit integers, defining a discrete set of quantization levels. The general quantization process involves scaling, rounding, and clamping, with variations depending on whether symmetric or asymmetric quantization is used.

{
In the following, we briefly explain the symmetric quantization and asymmetric quantization.
\begin{itemize}
    \item \textbf{Symmetric Quantization:}
    Symmetric quantization balances the quantization range around zero, ensuring that the positive and negative ranges are equal in magnitude. This is particularly useful in applications where data distributions (e.g., neural network weights) are centered around zero.

    For an $n$-bit signed integer in two's complement, the representable range is $[-2^{n-1}, 2^{n-1} - 1]$. The quantization range is defined symmetrically as $[-\omega, \omega]$, where $\omega$ is the maximum absolute value to be represented. The scaling factor $\Delta$ is:
    \begin{equation*}
    \Delta = \frac{\omega}{2^{n-1} - 1}.
    \end{equation*}

    The scaled value $s$ is computed as:
    \begin{equation*}
    s = \frac{r}{\Delta}.
    \end{equation*}

    The integer $x$ is obtained by rounding:
    \begin{equation*}
    \overline{x} = \mathrm{R}(s), \quad \text{ where }  \mathrm{R}(\cdot)\text{  denotes general rounding operator.}
    \end{equation*}

    Particularly, if $\mathrm{R}$ is round to nearest, ties to even, i.e., $\mathrm{R}  \equiv \mathrm{RNE}$, then $\mathrm{R}(s) = \lfloor s + 0.5 \rfloor$.

    Sometimes, a clamping is applied when needed, i.e.,
    \begin{equation*}
    x = \mathrm{max}(-2^{n-1}, \mathrm{min}(\overline{x}, 2^{n-1} - 1)).
    \end{equation*}

    To map $x$ back to $\hat{r}$, the dequantization is applied:
    \begin{equation*}
    \hat{r} = x \cdot \Delta.
    \end{equation*}
    Here, $r_{\text{min}} = -\omega$ and $r_{\text{max}} = \omega$, ensuring symmetry around zero. When $r = 0$, the quantized value is $x = 0$, preserving symmetry without an offset.

    \item \textbf{Asymmetric Quantization:}
    Asymmetric quantization allows unequal ranges for positive and negative values, typically by defining a range $[r_{\text{min}}, r_{\text{max}}]$ where $r_{\text{min}} \neq -r_{\text{max}}$. This is useful when the data distribution is skewed (e.g., all positive values in rectified activations like ReLU).

    For an $n$-bit signed integer, the range is still $[-2^{n-1}, 2^{n-1} - 1]$, but the quantization maps $[r_{\text{min}}, r_{\text{max}}]$ to this range. The scaling factor $\Delta$ is \begin{equation*}\Delta = \frac{r_{\text{max}} - r_{\text{min}}}{2^n - 1}.\end{equation*}

    The scaled value $s$ is
    \begin{equation*}
    s = \frac{r - r_{\text{min}}}{\Delta}.
    \end{equation*}
    Similarly, the integer $x$ is obtained by rounding and clamping:
    \begin{align*}
    \overline{x} &= \mathrm{R}(s), \quad \text{ where }  \mathrm{R}(\cdot)\text{  denotes general rounding operator,} \\
        x &= \mathrm{max}(-2^{n-1}, \mathrm{min}(\overline{x}, 2^{n-1} - 1)).
    \end{align*}

    Dequantization maps $x$ back to $\hat{r}$:
    \begin{equation*}
    \hat{r} = r_{\text{min}} + x \cdot \Delta.
    \end{equation*}
    In asymmetric quantization, the zero point (where $r = 0$) maps to an integer $z$ in the quantized domain:
    \begin{equation*}
    z = \mathrm{R}\left( \frac{0 - r_{\text{min}}}{\Delta} \right).
    \end{equation*}
    This introduces an offset, which may require additional computation during arithmetic operations.
\end{itemize}
}

Integer quantization {also} includes uniform quantization and non-uniform quantization. {Uniform quantization refers to dividing an integer range into equally-sized segments, while non-uniform quantization refers to using segments of varying sizes, often adjusted based on the integer values' distribution or importance}. Compared to non-uniform quantization, uniform quantization is computationally efficient, hardware-friendly, and easier to implement. Most modern processors, including CPUs, GPUs, and specialized accelerators like TPUs and NPUs, are optimized for integer arithmetic with uniform quantization, enabling fast matrix multiplications and reduced memory overhead.  While non-uniform quantization can provide better precision for highly skewed data distributions, it often requires more complex lookup tables or clustering methods, which increase computational cost and slow down inference. As a result, uniform quantization remains the standard choice for deep learning acceleration in both training and inference. Therefore, \texttt{pychop} focuses on uniform quantization.

In the following, we demonstrate the usage of \texttt{pychop} for integer quantization:
\begin{itemize}
    \item \begin{lstlisting}
Chopi(bits=8, symmetric=False, per_channel=False, axis=0)
\end{lstlisting}
The \texttt{Chopi} framework offers tailored integer quantization, allowing users to specify the bitwidth length (\texttt{bits}) and choose between symmetric or asymmetric quantization (\texttt{symmetric}). Additionally, two channel-specific parameters enable further customization: \texttt{per\_channel} determines whether quantization is applied on a per-channel basis, while \texttt{axis} specifies the dimension along which channel-wise quantization occurs when is set to \texttt{True}. A simple code demonstration for integer quantization is given in Appendix~\ref{app:iqt}.

\end{itemize}

\subsection{Common Mathematical Functions Support and Array Manipulation Routines}
We simulate common mathematical functions and operations (such as built-in functions in NumPy, PyTorch, or JAX) in reduced-precision arithmetic by first rounding the input to low precision, performing operations in the working precision, and then rounding the result back to low precision. This approach contrasts with CPFloat, which applies mathematical operations in working precision to inputs in working precision before rounding the final result to low precision.  A usage example is included in Appendix~\ref{app:math_func}.

\begin{table}[ht]\footnotesize
\centering
\caption{Function support and array manipulation routines (part I)}\footnotemark
\begin{tabular}{lp{9cm}}
\toprule
\textbf{Function} & \textbf{Description} \\
\addlinespace[0.5em]
\midrule
\rowcolor[gray]{0.9} \multicolumn{2}{l}{\textbf{Trigonometric Functions}} \\
\midrule
\texttt{sin} & Computes sine. Input in radians; output in $[-1, 1]$. \\
\texttt{cos} & Computes cosine. Input in radians; output in $[-1, 1]$. \\
\texttt{tan} & Computes tangent. Input in radians; discontinuities at $\pi/2 + k\pi$. \\
\texttt{arcsin} & Computes arcsin. Input in $[-1, 1]$; output in $[-\pi/2, \pi/2]$ radians. \\
\texttt{arccos} & Computes arccos. Input in $[-1, 1]$; output in $[0, \pi]$ radians. \\
\texttt{arctan} & Computes arctan. Output in $[-\pi/2, \pi/2]$ radians. \\
\addlinespace
\rowcolor[gray]{0.9} \multicolumn{2}{l}{\textbf{Hyperbolic Functions}} \\
\midrule
\texttt{sinh} & Computes hyperbolic sine. Output unrestricted. \\
\texttt{cosh} & Computes hyperbolic cosine. Output non-negative. \\
\texttt{tanh} & Computes hyperbolic tangent. Output in $(-1, 1)$. \\
\texttt{arcsinh} & Computes inverse hyperbolic sine. Output in real numbers. \\
\texttt{arccosh} & Computes inverse hyperbolic cosine. Input $\geq 1$; output in $[0, \infty)$. \\
\texttt{arctanh} & Computes inverse hyperbolic tangent. Input in $(-1, 1)$; output real. \\
\addlinespace
\rowcolor[gray]{0.9} \multicolumn{2}{l}{\textbf{Exponential and Logarithmic Functions}} \\
\midrule
\texttt{exp} & Computes $e^x$. Input unrestricted; output positive. \\
\texttt{expm1} & Computes $e^x - 1$. Enhanced precision for small $x$. \\
\texttt{log} & Computes natural logarithm (base $e$). Input positive; output unrestricted. \\
\texttt{log10} & Computes base-10 logarithm. Input positive; output unrestricted. \\
\texttt{log2} & Computes base-2 logarithm. Input positive; output unrestricted. \\
\texttt{log1p} & Computes $\log(1 + x)$. Input $> -1$; enhanced precision for small $x$. \\
\addlinespace
\rowcolor[gray]{0.9} \multicolumn{2}{l}{\textbf{Power and Root Functions}} \\
\midrule
\texttt{sqrt} & Computes square root. Input non-negative; output non-negative. \\
\texttt{cbrt} & Computes cube root. Input unrestricted; output sign matches input. \\
\addlinespace
\rowcolor[gray]{0.9} \multicolumn{2}{l}{\textbf{Aggregation and Linear Algebra Functions}} \\
\midrule
\texttt{sum} & Computes sum of array elements along axis. \\
\texttt{prod} & Computes product of array elements along axis. \\
\texttt{mean} & Computes mean of array elements along axis. \\
\texttt{std} & Computes standard deviation of array elements along axis. \\
\texttt{var} & Computes variance of array elements along axis. \\
\texttt{dot} & Computes dot product of two arrays. \\
\texttt{matmul} & Computes matrix multiplication of two arrays. \\
\addlinespace
\rowcolor[gray]{0.9} \multicolumn{2}{l}{\textbf{Special Functions}} \\
\midrule
\texttt{erf} & Computes error function. Output in $(-1, 1)$. \\
\texttt{erfc} & Computes complementary error function ($1 - \text{erf}$). \\
\texttt{gamma} & Computes gamma function. Input unrestricted. \\
\addlinespace
\rowcolor[gray]{0.9} \multicolumn{2}{l}{\textbf{Other Mathematical Functions}} \\
\midrule
\texttt{fabs} & Computes floating-point absolute value. Output non-negative. \\
\texttt{logaddexp} & Computes logarithm of sum of exponentials. \\
\texttt{cumsum} & Computes cumulative sum along axis. \\
\texttt{cumprod} & Computes cumulative product along axis. \\
\texttt{degrees} & Converts radians to degrees. \\
\texttt{radians} & Converts degrees to radians. \\
\addlinespace[0.5em]
\bottomrule
\end{tabular}
\end{table}

\begin{table}[ht]\footnotesize
\centering
\caption{Function support and array manipulation routines (part II)}\footnotemark
\begin{tabular}{lp{9cm}}
\toprule
\textbf{Function} & \textbf{Description} \\
\addlinespace[0.5em]
\midrule
\rowcolor[gray]{0.9} \multicolumn{2}{l}{\textbf{Rounding and Clipping Functions}} \\
\midrule
\texttt{floor} & Computes floor. Rounds down to nearest integer. \\
\texttt{ceil} & Computes ceiling. Rounds up to nearest integer. \\
\texttt{round} & Rounds to specified decimals. \\
\texttt{sign} & Computes sign. Returns $-1$, $0$, or $1$. \\
\texttt{clip} & Clips values to range $[a_{\text{min}}, a_{\text{max}}]$. \\
\addlinespace
\rowcolor[gray]{0.9} \multicolumn{2}{l}{\textbf{Miscellaneous Functions}} \\
\midrule
\texttt{abs} & Computes absolute value. Output non-negative. \\
\texttt{reciprocal} & Computes $1/x$. Input must not be zero. \\
\texttt{square} & Computes square of input. Output non-negative. \\
\addlinespace
\rowcolor[gray]{0.9} \multicolumn{2}{l}{\textbf{Additional Mathematical Functions}} \\
\midrule
\texttt{frexp} & Decomposes into significand and exponent. Chopping on significand. \\
\texttt{hypot} & Computes $\sqrt{x^2 + y^2}$. Inputs are real numbers. \\
\texttt{diff} & Computes difference between consecutive array elements. \\
\texttt{power} & Computes element-wise $x^y$. \\
\texttt{modf} & Decomposes into fractional and integral parts. Chopping on fractional part. \\
\texttt{ldexp} & Multiplies by $2$ to exponent power. \\
\texttt{angle} & Computes phase angle of complex number. Output in radians. \\
\texttt{real} & Extracts real part of complex number. \\
\texttt{imag} & Extracts imaginary part of complex number. \\
\texttt{conj} & Computes complex conjugate. \\
\texttt{maximum} & Computes element-wise maximum of two inputs. \\
\texttt{minimum} & Computes element-wise minimum of two inputs. \\
\addlinespace
\rowcolor[gray]{0.9} \multicolumn{2}{l}{\textbf{Binary Arithmetic Functions}} \\
\midrule
\texttt{multiply} & Computes element-wise product. \\
\texttt{mod} & Computes element-wise modulo. Divisor must not be zero. \\
\texttt{divide} & Computes element-wise division. Divisor must not be zero. \\
\texttt{add} & Computes element-wise sum. \\
\texttt{subtract} & Computes element-wise difference. \\
\texttt{floor\_divide} & Computes element-wise floor division. Divisor must not be zero. \\
\texttt{bitwise\_and} & Computes bitwise AND of integer inputs. \\
\texttt{bitwise\_or} & Computes bitwise OR of integer inputs. \\
\texttt{bitwise\_xor} & Computes bitwise XOR of integer inputs. \\
\addlinespace[0.5em]
\bottomrule
\end{tabular}
\end{table}

\footnotetext{All functions are computed with chopping to enforce reduced-precision format, where applicable.}

\subsection{Seamless NumPy / PyTorch / JAX  Integration}

\texttt{pychop} supports NumPy arrays, PyTorch tensors, and JAX arrays  as inputs for computations on their respective backends. Each backend brings its own advantages: NumPy excels in straightforward CPU vectorized computation and has a broader range of applications for scientific computing; PyTorch's primary advantages lie in GPU acceleration and dynamic computation graphs (eager execution), making \texttt{pychop} particularly suitable for deep learning training and large-batch tensor operations; JAX leverages just-in-time (JIT) compilation via XLA and its functional pure-function characteristics, achieving exceptionally high post-compilation performance on CPU, GPU, and TPU. The default backend is \texttt{auto}: \texttt{pychop} detects supported input types when a quantizer is called, while users may also manually specify a backend to avoid repeated detection or to target a specific framework. Optional deep-learning backends are imported lazily, so importing \texttt{pychop} does not eagerly import PyTorch, or JAX. The associated code example is explained in Appendix~\ref{app:backend}.

\subsection{Neural Network Quantization}

Mixed-precision Deep Neural Networks provide the energy efficiency and throughput essential for hardware deployment, particularly in resource-limited settings, often without compromising accuracy. However, identifying the optimal per-layer bit precision remains challenging due to the vast search space introduced by the diverse range of models, datasets, and quantization techniques (see \cite{10509805} and references therein). Neural network training and inference are inherently resilient to errors, a characteristic that distinguishes them from traditional workloads that demand precise computations and high dynamic range number representations. It is well understood that, given the presence of statistical approximation and estimation errors, high-precision computations in learning tasks are often unnecessary \citep{NIPS2007_0d3180d6}. Furthermore, introducing noise during training has been shown to improve neural network performance \citep{6796505, 6707022, KOSKO2020359}.

\texttt{pychop} is well-suited for neural network deployment with post-training quantization (PTQ) and quantization-aware training (QAT). Its design prioritizes simplicity and flexibility, making it an ideal tool for experimenting with and fine-tuning quantization strategies. In the following, we provide a concise illustration of how \texttt{pychop} can be effectively utilized in quantization applications for neural networks, demonstrating its ease of use and integration into existing workflows. This process is adapted as follows:
\begin{itemize}
    \item \textbf{Training}: During quantization-aware training (QAT), the network simulates fixed-point arithmetic by quantizing weights and activations in the forward pass. Gradients may remain in higher precision.
    \item \textbf{Inference}: Weights and activations are quantized to required format for efficient computation.
\end{itemize}

\paragraph{Principle and Basic Usage}

\begin{table}[ht]
\centering\small
\caption{Commonly implemented quantized layers (part) and their original PyTorch names. Layers prefixed with “Quantized” refer to layers designed for floating-point and fixed-point quantization, while layers prefixed with “IQuantized” refer to layers designed for integer quantization.}\label{tab:layers}
\begin{tabular}{|c|c|}
\hline
\textbf{Quantized Layer Name} & \textbf{Original PyTorch Name} \\
\hline
QuantizedLinear /  IQuantizedLinear & nn.Linear \\
QuantizedConv1d / IQuantizedConv1d & nn.Conv1d \\
QuantizedConv2d /  IQuantizedConv2d & nn.Conv2d \\
QuantizedConv3d /  IQuantizedConv3d  & nn.Conv3d \\
QuantizedRNN / IQuantizedRNN& nn.RNN \\
QuantizedLSTM / IQuantizedLSTM & nn.LSTM \\
QuantizedMaxPool1d / IQuantizedMaxPool1d& nn.MaxPool1d \\
QuantizedMaxPool2d / IQuantizedMaxPool2d & nn.MaxPool2d \\
QuantizedMaxPool3d / IQuantizedMaxPool3d & nn.MaxPool3d \\
QuantizedAvgPool1d/2d/3d / IQuantizedAvgPool1d/2d/3d & nn.AvgPool1d/2d/3d \\
QuantizedMultiheadAttention / IQuantizedMultiheadAttention & nn.MultiheadAttention \\
QuantizedBatchNorm1d / IQuantizedBatchNorm1d & nn.BatchNorm1d \\
QuantizedBatchNorm2d / IQuantizedBatchNorm2d & nn.BatchNorm2d \\
QuantizedBatchNorm3d /  IQuantizedBatchNorm3d & nn.BatchNorm3d \\
\hline
\end{tabular}
\end{table}

\begin{table}[ht]
\centering\small
\caption{Quantized optimizers (part) and their original PyTorch names. Optimizers prefixed with “Quantized” refer to floating-point and fixed-point optimizer-state quantization, while optimizers prefixed with “IQuantized” refer to integer quantization.}\label{tab:optim}
\begin{tabular}{|c|c|}
\hline
\textbf{Common quantized optimizer name} & \textbf{Original PyTorch name} \\
\hline
QuantizedSGD / IQuantizedSGD& torch.optim.SGD \\
QuantizedAdam /  IQuantizedAdam& torch.optim.Adam \\
QuantizedRMSprop / IQuantizedRMSprop & torch.optim.RMSprop \\
QuantizedAdagrad /  IQuantizedAdagrad & torch.optim.Adagrad \\
QuantizedAdadelta /  IQuantizedAdadelta & torch.optim.Adadelta \\
QuantizedAdamW / IQuantizedAdamW & torch.optim.AdamW \\
\hline
\end{tabular}
\end{table}

\texttt{pychop} simulates multiple-precision neural network training by introducing floating-point, fixed-point, or integer quantization into the training process while still performing the underlying computations in the host framework's working precision. The examples and tables below focus on the PyTorch interface, while the package also provides JAX/Flax interfaces. The pre-built quantized layers and optimizer classes extend the multiple-precision emulation of \texttt{torch.nn.Module} and algorithms in \texttt{torch.optim}, applying the simulator to various layer and arithmetic operations. Part of the implemented quantized layers and optimization algorithms are listed in \tablename~\ref{tab:layers} and \tablename~\ref{tab:optim}. All layers and optimizers follow a modular design for easy extension, with the same parameter settings of the original PyTorch modules plus additional parameters to define rounding modes and quantization settings such as bitwidth for exponent and significand (\texttt{exp\_bits} and \texttt{sig\_bits}), and preserve original tensor shapes and PyTorch compatibility. As for optimizers, the quantization can be applied to gradients, momenta, and other state variables used by the optimizers. The design of this functionality facilitates the study of quantization effects in neural network performance, the simulation of reduced-precision hardware, and the evaluation of numerical stability in deep learning. We briefly summarize these functions as follows:
\begin{itemize}
\item \textbf{Implementation}: For layers, \texttt{pychop} quantizes weights, input, and bias before operations, then uses standard PyTorch matrix multiplication and addition with working precision (either fp32 or fp64, which depends on user settings). The gradient flow through the quantized operations is maintained in working precision.

\item \textbf{Parameters}: \texttt{pychop} allows the quantization of weights and biases during initialization or forward pass, and quantizes inputs, performs matrix multiplication, and adds quantized bias, all in the specified format.
\item \textbf{Flexibility}: \texttt{pychop} allows the quantization of different parts of the training process independently, such as weights, activations, gradients, momentum, and gradient accumulators. It also provides the pre-built layers and optimizers for training.
It supports customizable reduced-precision formats, including floating-point (with configurable bitwidth for exponent and significand parts), fixed-point (with configurable bitwidth for integer and fraction parts), and integers arithmetic (with configurable bitwidth for integer part). % For example, you can define an 8-bit floating-point format with 5 exponent bits and 2 significand bits.

\item \textbf{Extensibility}: Template design supports adaptation to convolutional or recurrent layers. It also provides built-in quantized layers, for example, \texttt{QuantizedLinear}, \texttt{QuantizedRNN}, \texttt{QuantizedLSTM}, \texttt{QuantizedGRU}, which corresponds to the reduced-precision emulation of \texttt{nn.Linear}, \texttt{nn.RNN}, \texttt{nn.LSTM}, and \texttt{nn.GRU}.  The example is illustrated in Appendix~\ref{app:qat}.
\end{itemize}

\texttt{pychop} further provides the interface \texttt{pychop.layers.post\_quantization} to convert model parameters into customized precision by specifying rounding and format parameters, enabling post-quantization emulation. We demonstrate the usage as below.
\begin{lstlisting}[title={Post quantization for neural network deployment.}]
from pychop import Chop
from pychop.layers import post_quantization

quantizer = Chop(exp_bits=5, sig_bits=10, rmode=1)
quantized_model = post_quantization(model, quantizer)
\end{lstlisting}

\paragraph{Straight-Through Estimator}
The Straight-Through Estimator (STE) is a methodological framework widely used in training neural networks with discrete or non-differentiable operations, such as quantization or binarization. These operations challenge conventional backpropagation, which requires continuous gradients for parameter optimization. Non-differentiable functions, with their zero or undefined gradients, obstruct this process, impeding effective learning. The STE addresses this by approximating the gradient to enable training despite such discontinuities.

The STE operates by treating a non-differentiable function as differentiable during backward propagation. In the forward pass, it applies the intended discrete transformation, such as rounding a continuous value. In the backward pass, rather than using the operation’s true gradient—typically zero or undefined—it directly propagates the gradient from subsequent layers to preceding ones, bypassing the discrete step. This approximation allows gradient-based optimization to proceed, expanding the range of trainable neural architectures.

The \texttt{pychop} framework integrates STE modules to support quantization-aware training. Specifically, \texttt{ChopSTE}, \texttt{ChopfSTE}, \texttt{ChopiSTE}, and the quantized layer wrappers apply the intended floating-point, fixed-point, or integer quantization in the forward pass while permitting gradients to propagate through during the backward pass as if the quantization operation had not occurred. This approach effectively reconciles the challenges posed by non-differentiable quantization, ensuring robust training of quantized neural networks.

\subsection{Support for MATLAB}

MATLAB provides built-in support for calling Python libraries through its Python Interface. This allows users to use Python functions, classes, and modules directly from MATLAB, making it easy to integrate Python-based scientific computing, machine learning, and deep learning libraries into MATLAB workflows. MATLAB interacts with Python by adding the \texttt{py.} prefix, which allows MATLAB to call the needed Python library seamlessly. One can also execute Python statements in the Python interpreter directly from MATLAB using the \texttt{pyrun} or \texttt{pyrunfile} functions. The setup is illustrated in Appendix~\ref{app:matlab}.
For details, we refer users to \url{https://www.mathworks.com/help/matlab/call-python-libraries.html}.

\section{Simulations}\label{sec:exps}

\subsection{Environmental Settings}
Our experiments are simulated on a Dell PowerEdge R750xa server\footnote{\url{https://front.convergence.lip6.fr/}} with 2 TB of memory, Intel Xeon Gold 6330 processors (56 cores, 112 threads, 2.00 GHz), and an NVIDIA A100 GPU (80 GB HBM2, PCIe), providing robust computational power for large-scale simulations and deep learning tasks. We simulate the code in Python 3.10 and MATLAB R2024b. {All simulations are performed on a single CPU and GPU}.  In all experiments involving runtime measurements, we perform multiple runs, discard the first warm-up run, and record the average of the remaining runs as the measured runtime. In the MATLAB simulation, we run the experiment 11 times and take the average of the last 10 runs, whereas for other experiments we run 4 times and take the average of the last 3 runs. We perform more runs in the MATLAB simulation to better smooth out outliers arising from function calls from Python within MATLAB.  
The experiments in this section use \texttt{pychop} version 0.6.0.

We simulated reduced-precision quantization using \texttt{pychop} on a variety of benchmark datasets intended for a broad range of computer vision tasks to evaluate the effect of reduced-precision quantization on object recognition and image classification:
\begin{itemize}
\item \textbf{MNIST}~\citep{6296535}: The dataset comprises 60,000 training and 10,000 test grayscale images of handwritten digits (0--9), each with a resolution of $28 \times 28$ pixels. Widely used as a benchmark for image classification and optical character recognition, the images are preprocessed to be centered and normalized for consistent sizing and intensity.
\item \textbf{Fashion-MNIST}~\citep{xiao2017fashion}: The dataset contains 60,000 training and 10,000 test grayscale images of fashion items from Zalando’s inventory, each with a resolution of $28 \times 28$ pixels. Spanning 10 classes (e.g., clothing, shoes, bags), it serves as a more complex alternative to MNIST for benchmarking image classification models.
\item \textbf{Caltech101}~\citep{ranzato_perona_2022}: The dataset comprises approximately 9,144 RGB images of objects across 101 categories (e.g., animals, vehicles, household items) and an additional background class. Image sizes vary, typically around 300 pixels on the longer side, and are often resized (e.g., to $224 \times 224$) for specific tasks. With imbalanced sample sizes (40--800 images per category), it provides a challenging benchmark for image classification due to its diverse and heterogeneous visual patterns.
\item \textbf{Oxford-IIIT Pet}~\citep{10.5555/2354409.2355061}: The dataset contains approximately 7,349 RGB images of cats and dogs across 37 breeds (12 cats, 25 dogs), with about 200 images per breed, split into training/validation (3,680 images) and test (3,669 images) sets. This dataset supports fine-grained classification, with significant variability in pose, lighting, and background, making it suitable for real-world visual discrimination tasks. Here, images are
resized to $256 \times 256$ followed by a crop to $224 \times 224$ for analysis.

\item \textbf{COCO}~\citep{lin2015miccoco}: The dataset contains RGB images across 80 categories (e.g., people, animals, vehicles, household items), designed for object detection, segmentation, and captioning. It features complex backgrounds, multiple objects per image, and annotations for bounding boxes and segmentation masks. For our simulation, we used the COCO val2017 subset ($\sim$5,000 images) to efficiently evaluate our quantized Faster R-CNN model on a diverse, well-annotated set, avoiding the computational cost of the full training set ($\sim$118,000 images). Here, images are resized to $256 \times 256$ followed by a crop to $224 \times 224$ for analysis.

\end{itemize}

\subsection{Speedup in MATLAB}

% Since MATLAB provides Python virtual environment which enable accessibility of the Python code in MATLAB
Experimental simulations were conducted to compare the runtime performance of MATLAB's \texttt{chop} function with \texttt{pychop} for half-precision and bfloat16 precision rounding within the MATLAB environment. Additionally, \texttt{pychop}'s performance was independently evaluated in a Python environment across various computational frameworks (NumPy and PyTorch) and hardware configurations (on CPU and GPU). The study assessed the baseline performance of MATLAB's \texttt{chop} alongside \texttt{pychop}, which implements the \texttt{Chop} and \texttt{FaultChop} methods. Simulations tested {square} matrix sizes of $2{,}000$, $4{,}000$, $6{,}000$, $8{,}000$, and $10{,}000$, where elements were randomly generated in double-precision and uniformly distributed in $[-1, 1]$. We employ multiple rounding modes: round to nearest, round up, round down, round toward zero, and stochastic rounding. For clarity in the following discussion, \texttt{pychop}'s \texttt{Chop} and \texttt{FaultChop} are referred to simply as \texttt{Chop} and \texttt{FaultChop}, respectively.

\begin{figure}[ht]
\includegraphics[width=14.3cm]{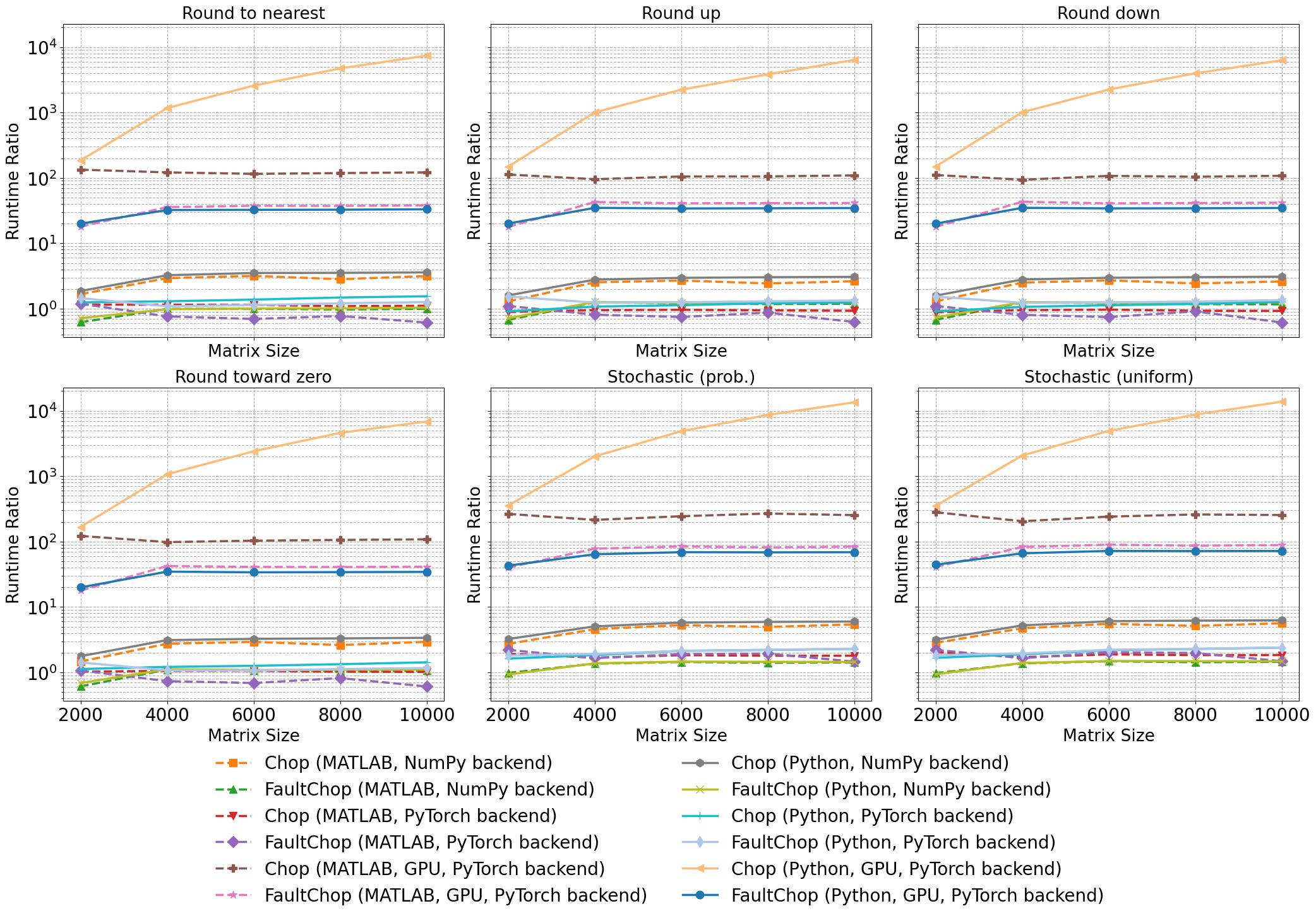}
\caption{Runtime ratio of MATLAB's \texttt{chop} over \texttt{pychop} in half precision (dashed for MATLAB-based, solid for Python-based).}\label{fig:matlab1}
\end{figure}

\begin{figure}[ht]
\includegraphics[width=14.3cm]{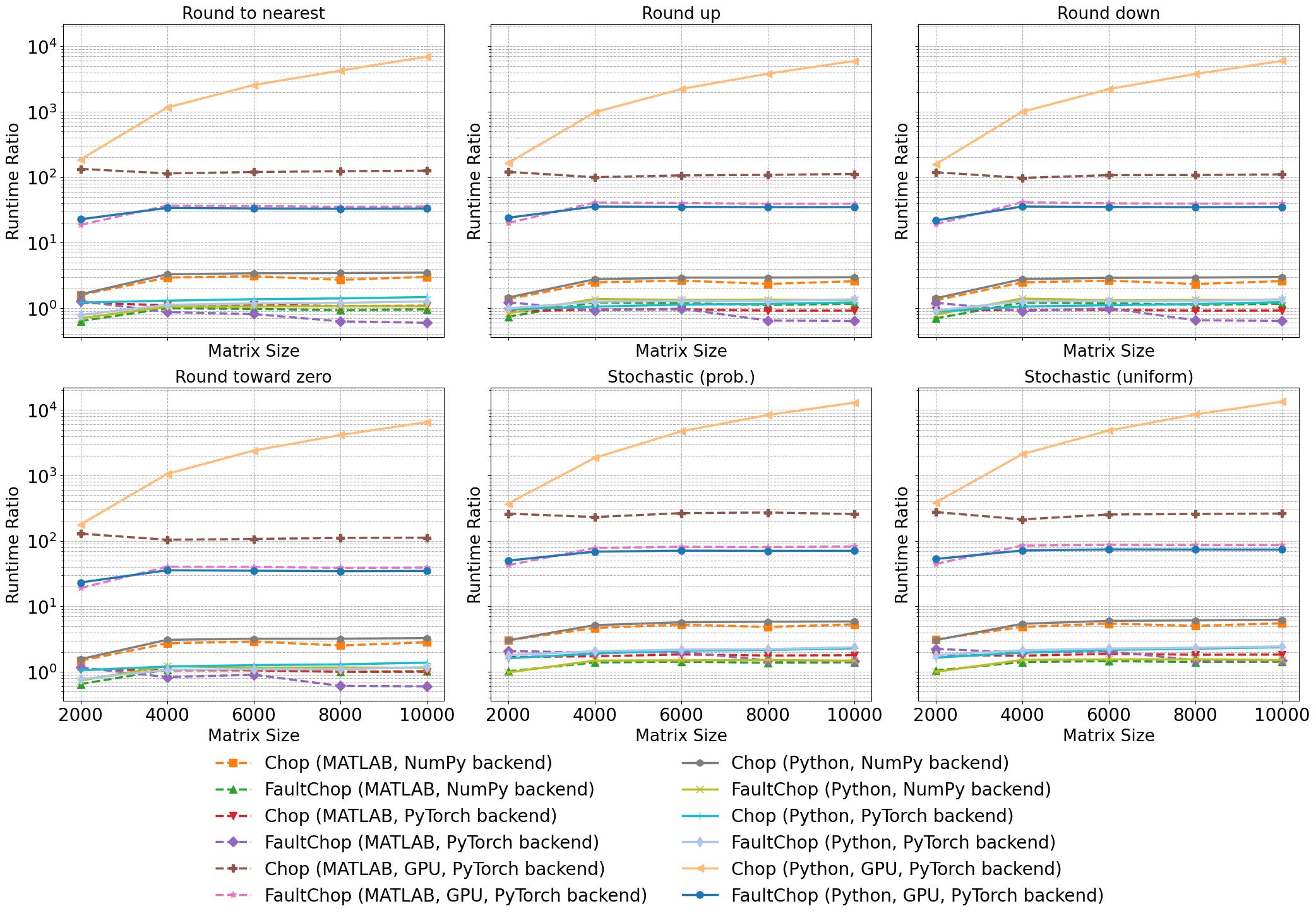}
\caption{Runtime ratio of MATLAB's \texttt{chop} over \texttt{pychop} in bf16 precision (dashed for MATLAB-based, solid for Python-based).}\label{fig:matlab2}
\end{figure}

The \figurename~\ref{fig:matlab1} and~\ref{fig:matlab2} illustrate the runtime performance of the baseline MATLAB's \texttt{Chop} in comparison to \texttt{pychop}, offering insights into scalability trends, framework efficiency, hardware influences, and optimization benefits. Results are presented in semilogarithmic plots, with distinct line styles distinguishing the data.

Although invoking \texttt{pychop} within MATLAB introduces some runtime overhead, \texttt{Chop} consistently outperforms MATLAB's \texttt{Chop}, while \texttt{FaultChop} on the CPU exhibits performance comparable to MATLAB's \texttt{Chop}. Furthermore, both \texttt{Chop} and \texttt{FaultChop} achieve speedups of orders of magnitude over MATLAB's \texttt{Chop} when deployed on GPU hardware. Notably, the speedup ratio of \texttt{Chop} becomes increasingly pronounced as the matrix size grows. Our results also show that stochastic rounding modes achieve performance comparable to deterministic modes on all backends.

\subsection{Backend-Specific Performance Breakdown}

\begin{figure}[htp]
\centering
\includegraphics[width=0.39\linewidth]{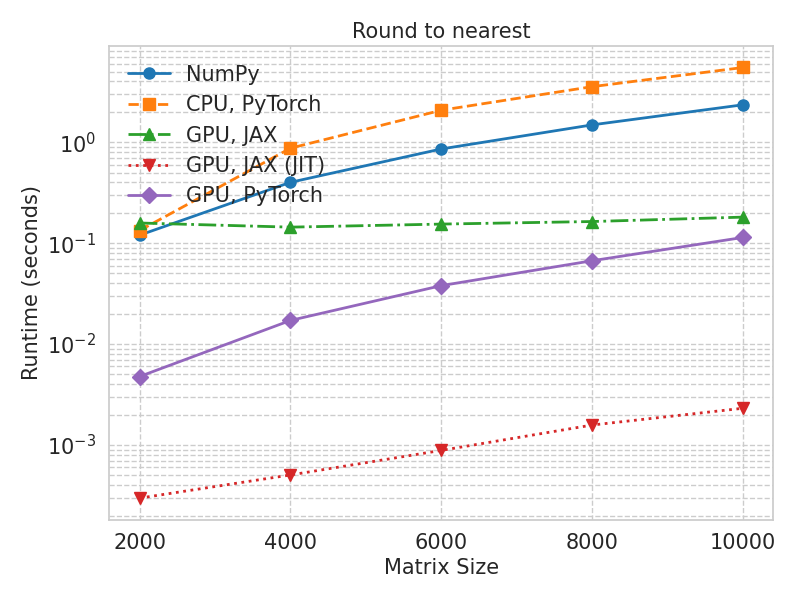}
\includegraphics[width=0.39\linewidth]{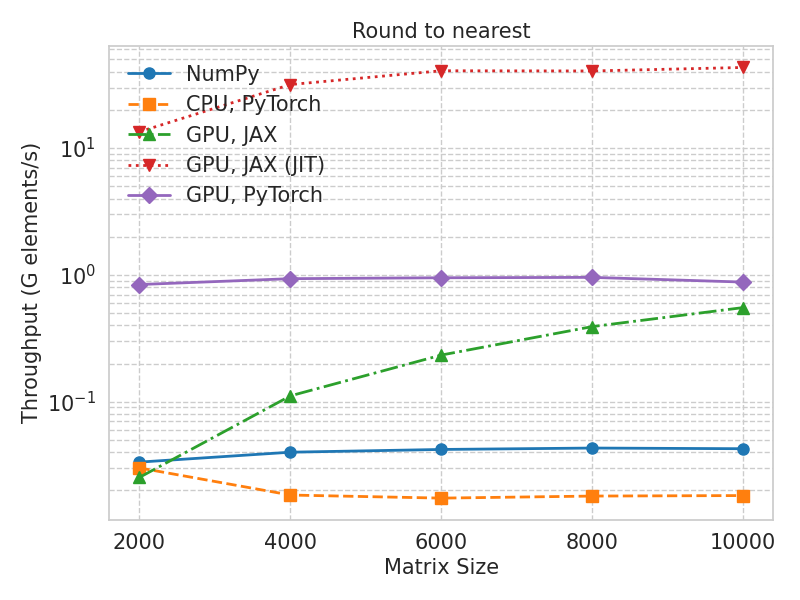}
\includegraphics[width=0.39\linewidth]{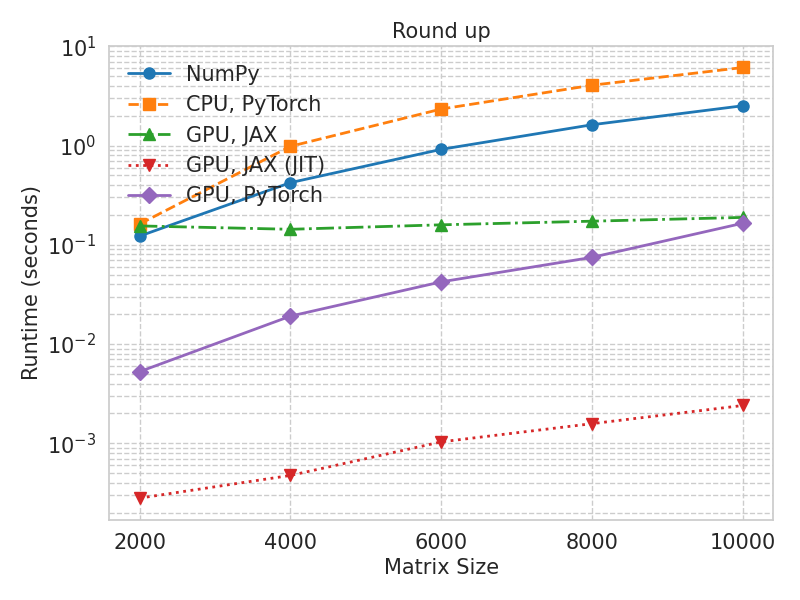}
\includegraphics[width=0.39\linewidth]{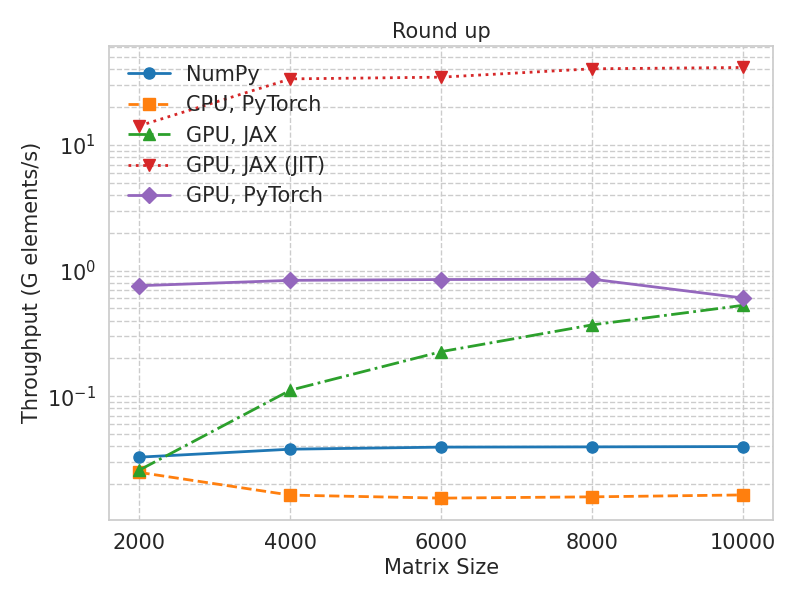}
\includegraphics[width=0.39\linewidth]{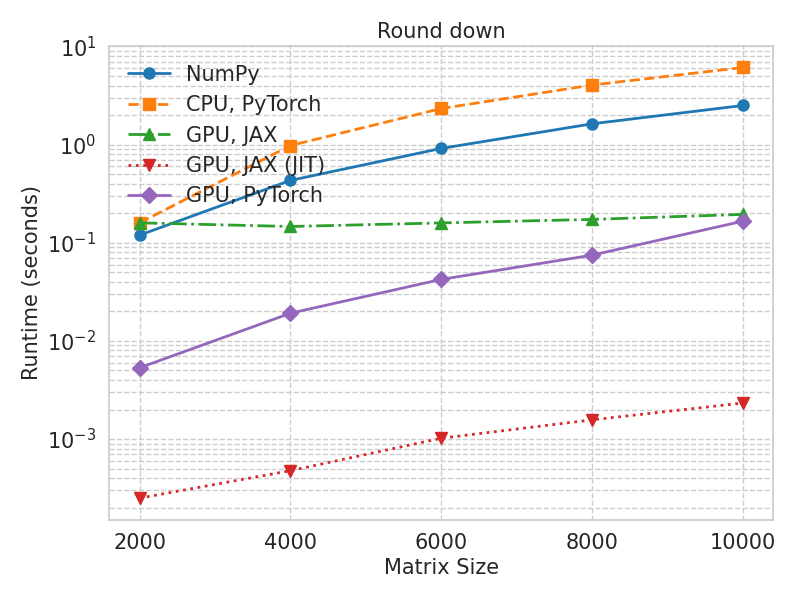}
\includegraphics[width=0.39\linewidth]{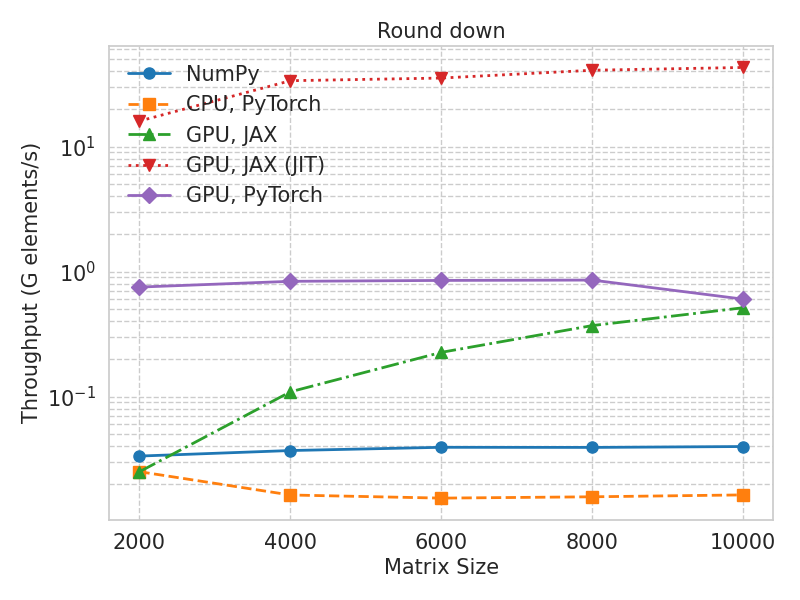}
\includegraphics[width=0.39\linewidth]{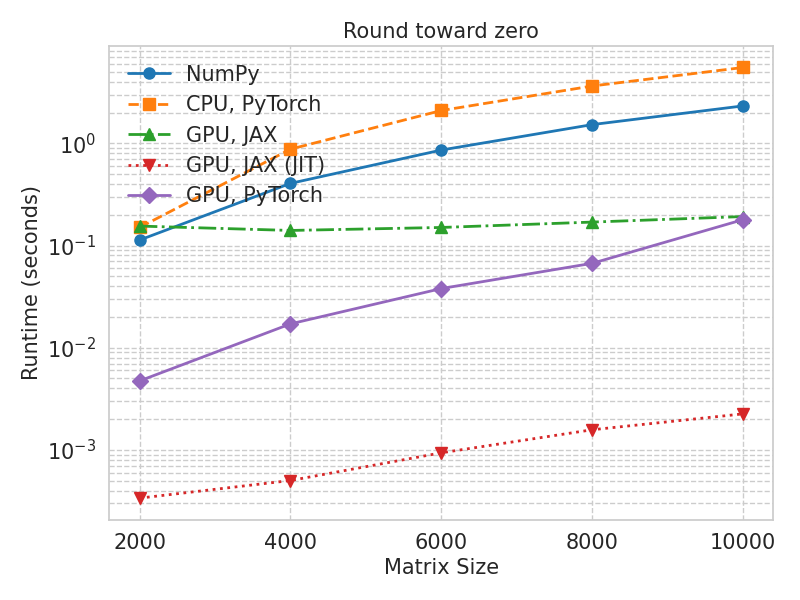}
\includegraphics[width=0.39\linewidth]{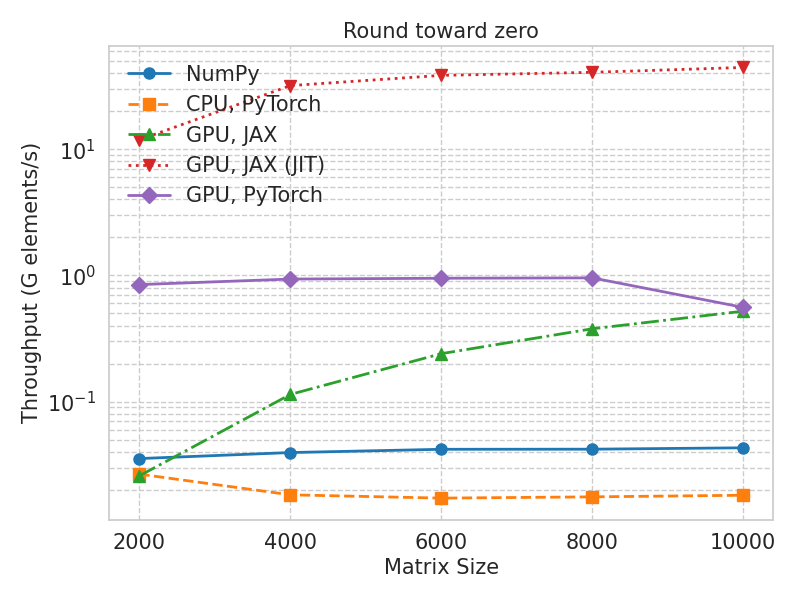}
\caption{Runtime and throughput of quantize-only operation with deterministic rounding.}
\label{fig:runtime-and-throughput1}
\end{figure}

We also conducted additional experiments to provide a detailed breakdown of runtime and throughput across the different backends supported by \texttt{Chop}: NumPy, PyTorch (CPU and GPU), and JAX (eager and JIT-compiled modes). Similarly to the last experiment, we evaluated the quantize-only operation on square matrices sizes of 2000, 4000, 6000, 8000, and 10000. We quantized to bf16 precision. All six supported rounding modes were tested: round to nearest, round up, round down, round
toward zero, and stochastic rounding (proportional and uniform). Each configuration was run 4 times, discarding the first run as warmup to account for JIT compilation and caching effects. Runtimes were measured using wall-clock time, and throughput $\tau$, i.e., the number of elements processed per second (reported in Giga-elements/s), was computed as
\begin{equation*}
    \tau = \frac{N^2}{t \cdot 10^9},
\end{equation*}
in which $N$ and $t$ are referred to as matrix sizes and wall-clock time, respectively.  Experiments were conducted with memory cleanup performed between runs to ensure consistent conditions.

\begin{figure}[htp]
    \centering
    \includegraphics[width=0.39\linewidth]{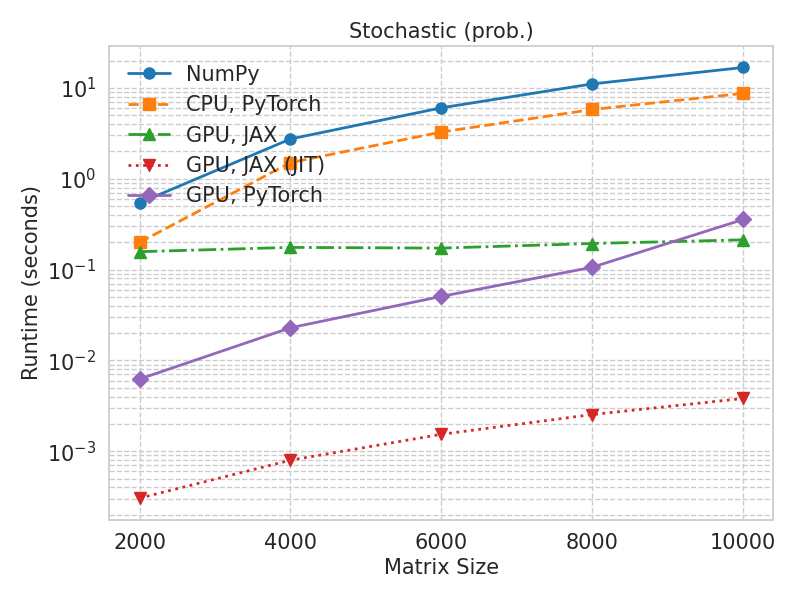}
    \includegraphics[width=0.39\linewidth]{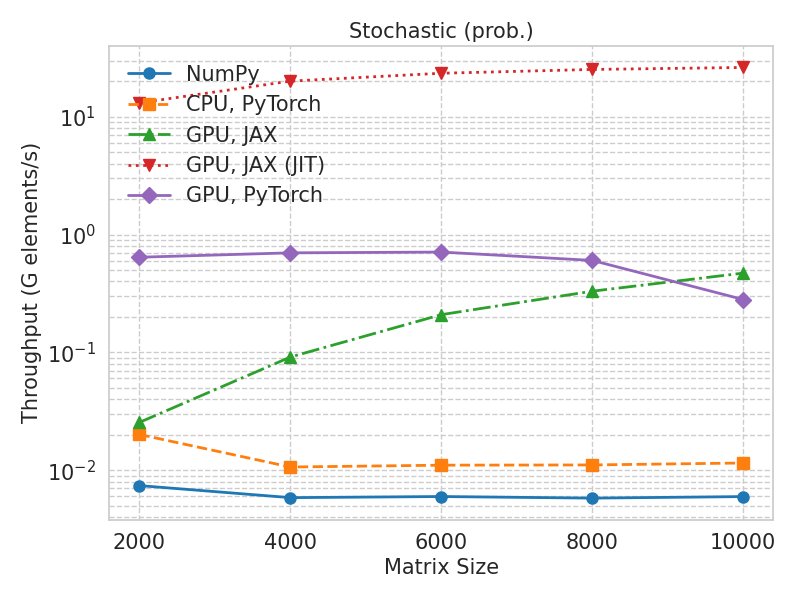}
    \includegraphics[width=0.39\linewidth]{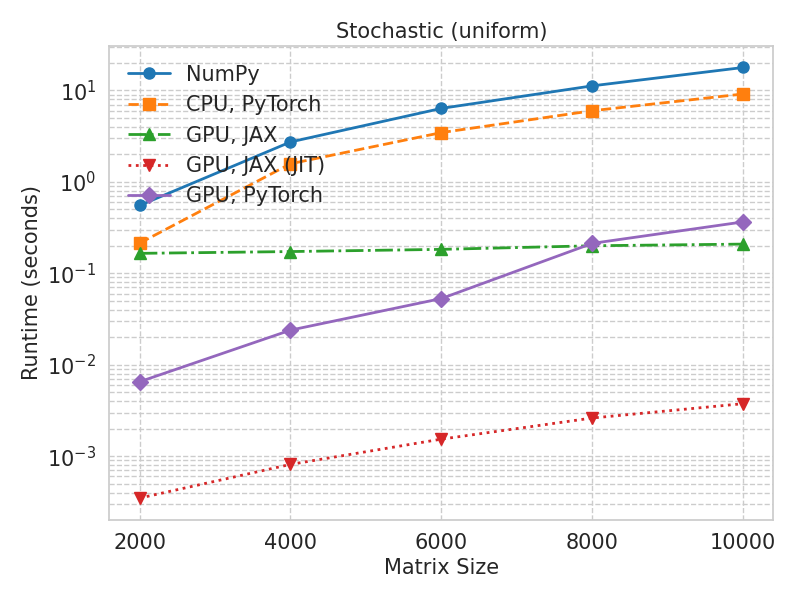}
    \includegraphics[width=0.39\linewidth]{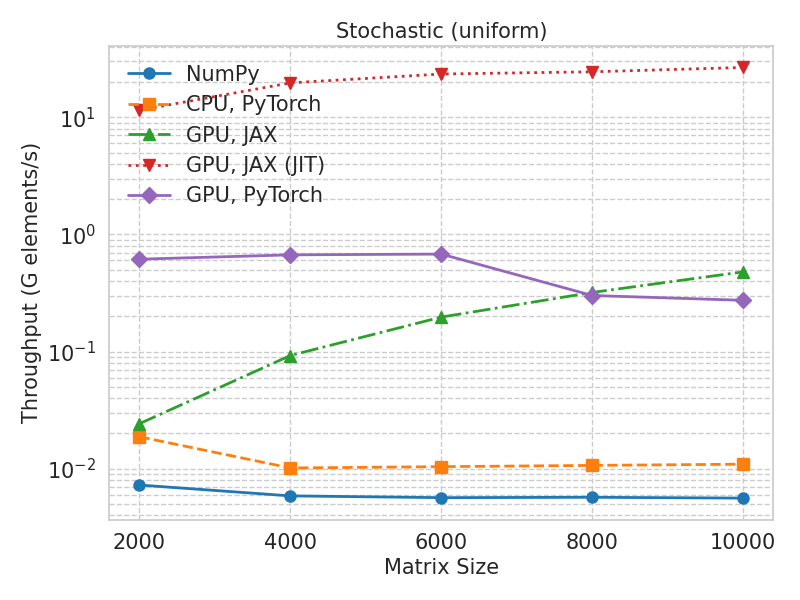}
    \caption{Runtime and throughput of quantize-only operation with stochastic rounding.}
    \label{fig:runtime-and-throughput2}
\end{figure}

% clean up = \texttt{gc.collect()} and \texttt{torch.cuda.empty\_cache()}

Figures~\ref{fig:runtime-and-throughput1} and~\ref{fig:runtime-and-throughput2} depict runtime and throughput across matrix sizes and rounding modes. All rounding modes exhibit similar patterns regarding speed and throughput.   On GPU, the JAX backend with JIT run on GPU achieves substantially higher performance, up to orders of magnitude, than the NumPy backend, demonstrating effective utilization of massive parallelism for element-wise quantization. Without JIT, JAX performs slightly slower than Torch on GPU, however, the performance gap narrows as the number of elements for quantization grows, where launch overhead and memory bandwidth dominate.  Additionally, the results show that the performance of stochastic rounding is similar to the performance of deterministic modes across all backends.

\begin{figure}
    \centering
    \includegraphics[width=0.76\linewidth]{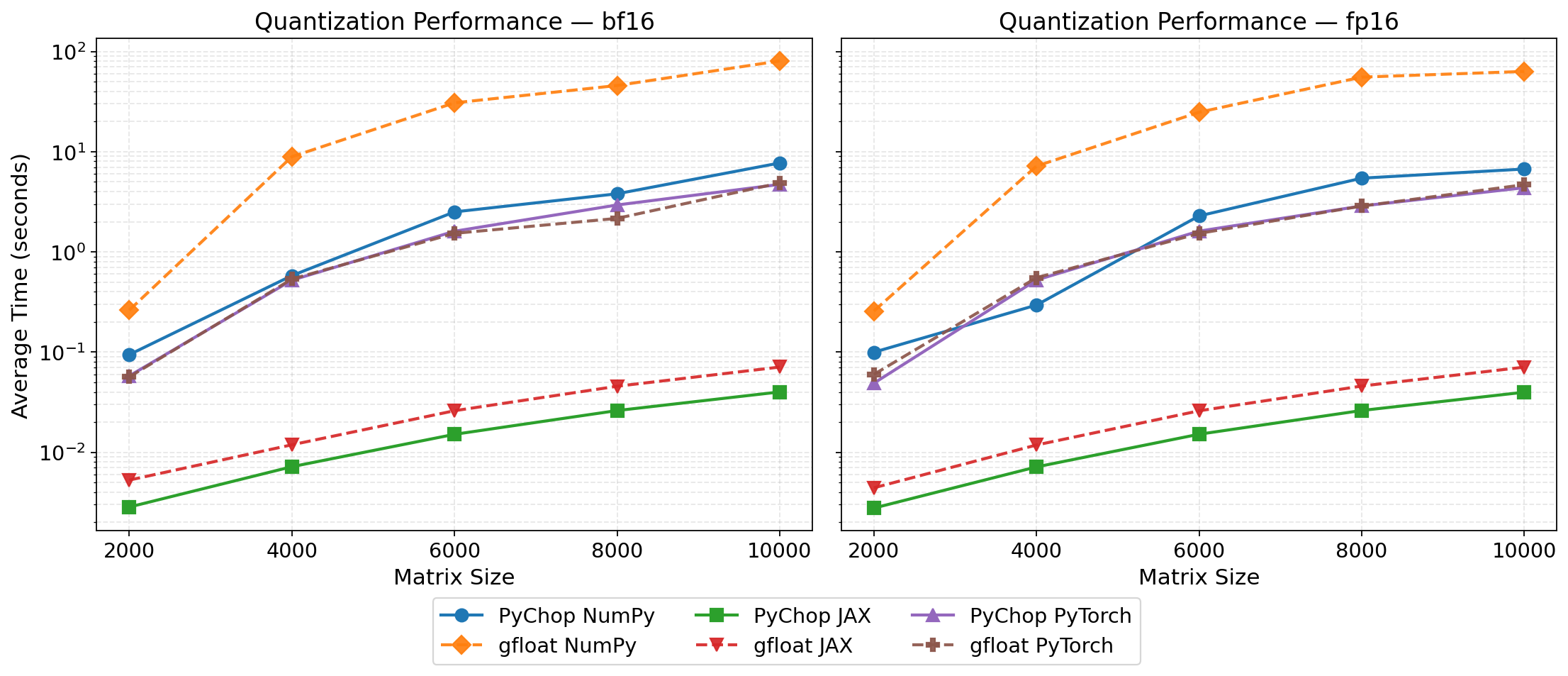}
    \caption{Runtime comparison between \texttt{pychop} and \texttt{gfloat}.}
    \label{fig:gfloat_comparison}
\end{figure}

To further demonstrate the performance across backends, we compare \texttt{pychop} with \texttt{gfloat} using the same matrix sizes, for matrices with elements generated using the standard normal distribution. We feed the two libraries with array type in NumPy, PyTorch, and JAX, respectively, and specify the corresponding backend for \texttt{pychop}. Except for NumPy, the computation in other backends were performed in GPU run. As shown in \figurename~\ref{fig:gfloat_comparison}, across different matrix sizes, \texttt{pychop} and \texttt{gfloat} achieve similar performance for PyTorch arrays, while \texttt{pychop} outperforms \texttt{gfloat} by orders of magnitude with both NumPy and JAX arrays;  the pronounced outperforming of \texttt{pychop} occurs particularly in NumPy.

\subsection{Extra overhead from Precision Emulation}

In this subsection, we demonstrate how much overhead is incurred by using \texttt{pychop} for precision emulation in numerical methods. We compared iterative refinement in uniform precision fp32 with both emulated fp32 and native fp32. Here we solve linear systems $Ax = b$, where $A$ is a nonsymmetric, positive definite matrix with a 2-norm medium condition number of approximately $10^{4}$. The iterative refinement performs LU factorization once in fp32 followed by repeated residual computation and correction using the same factors. The eventual fp32 solution and approximate solution agree within a relative tolerance of $10^{-8}$. The iterative refinement follows the implementation of \cite{higham2002accuracy}.

\begin{figure}[htp]
\centering
\subfigure[Matrix size=$2000$]{\includegraphics[width=0.35\linewidth]{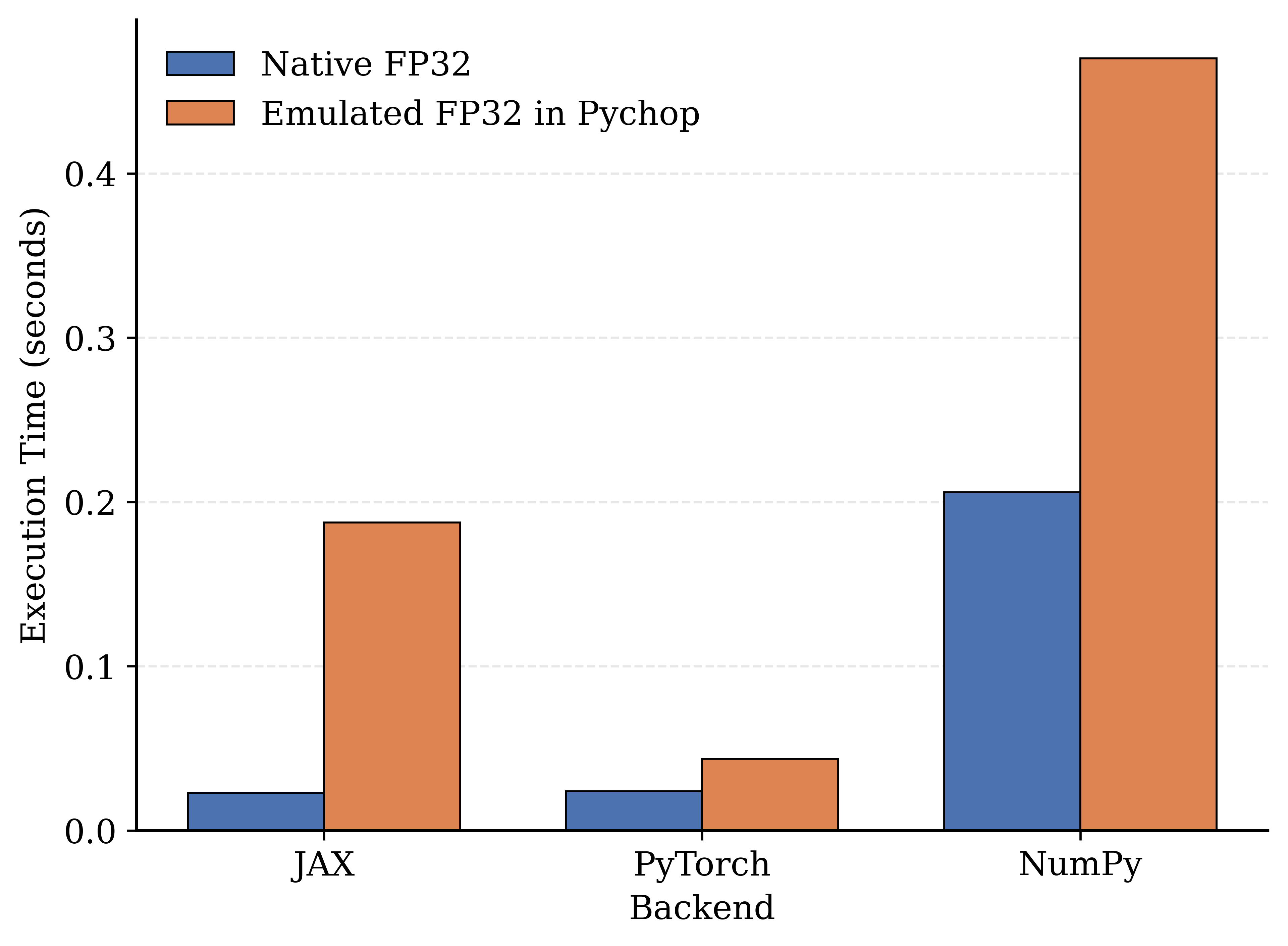}}
\subfigure[Matrix size=$4000$]{\includegraphics[width=0.35\linewidth]{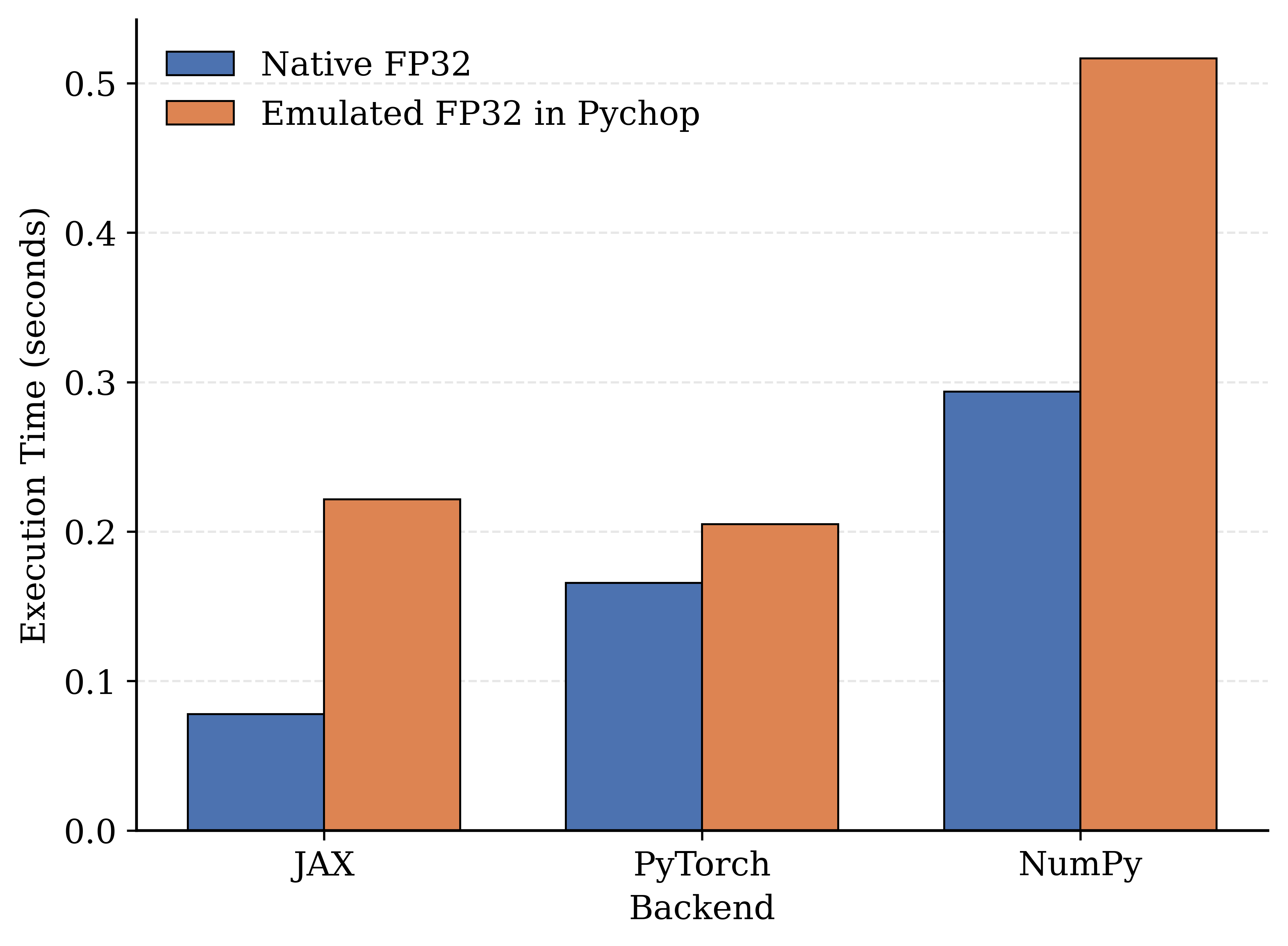}}
\subfigure[Matrix size=$6000$]{\includegraphics[width=0.35\linewidth]{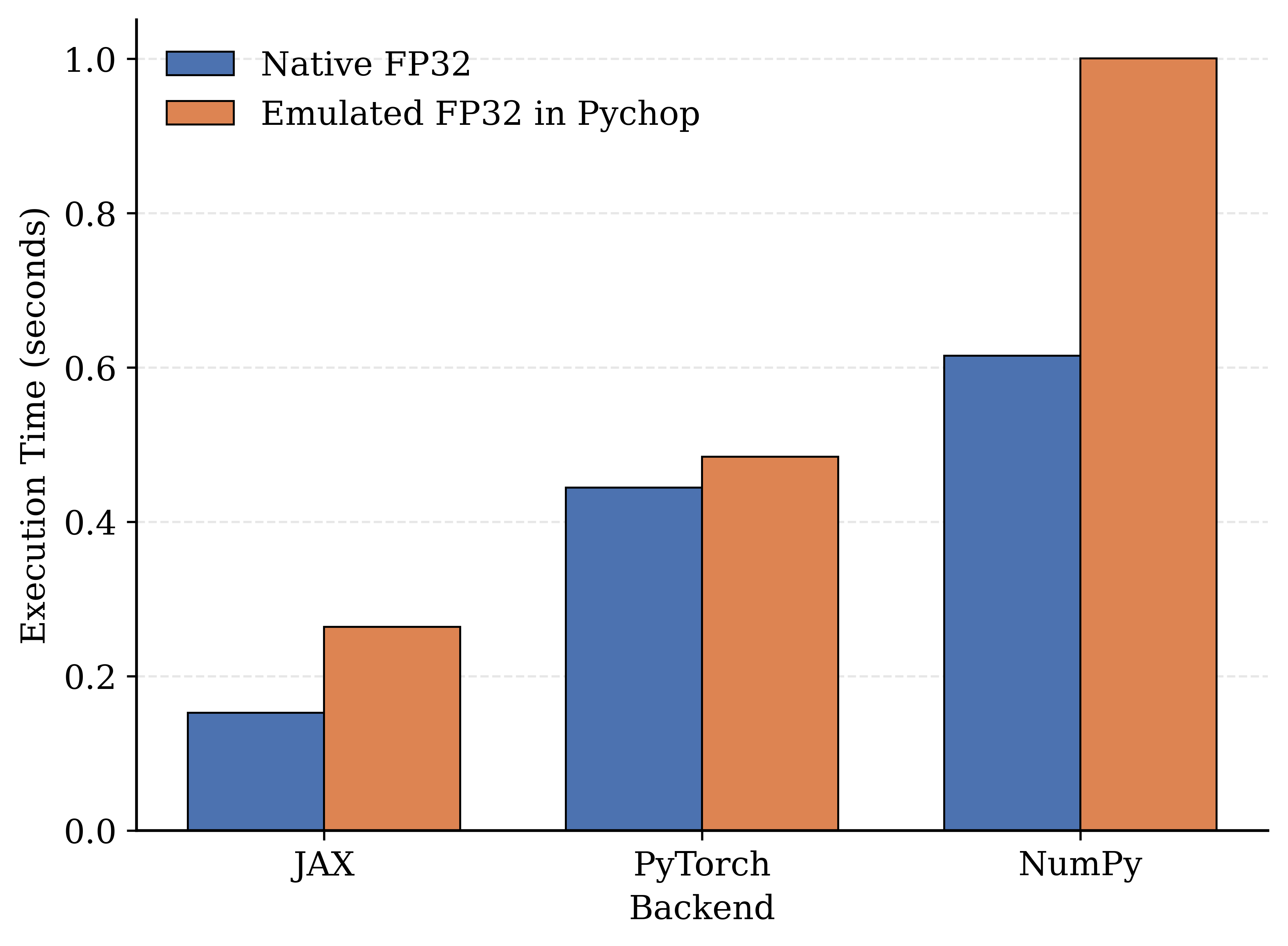}}
\subfigure[Matrix size=$8000$]{\includegraphics[width=0.35\linewidth]{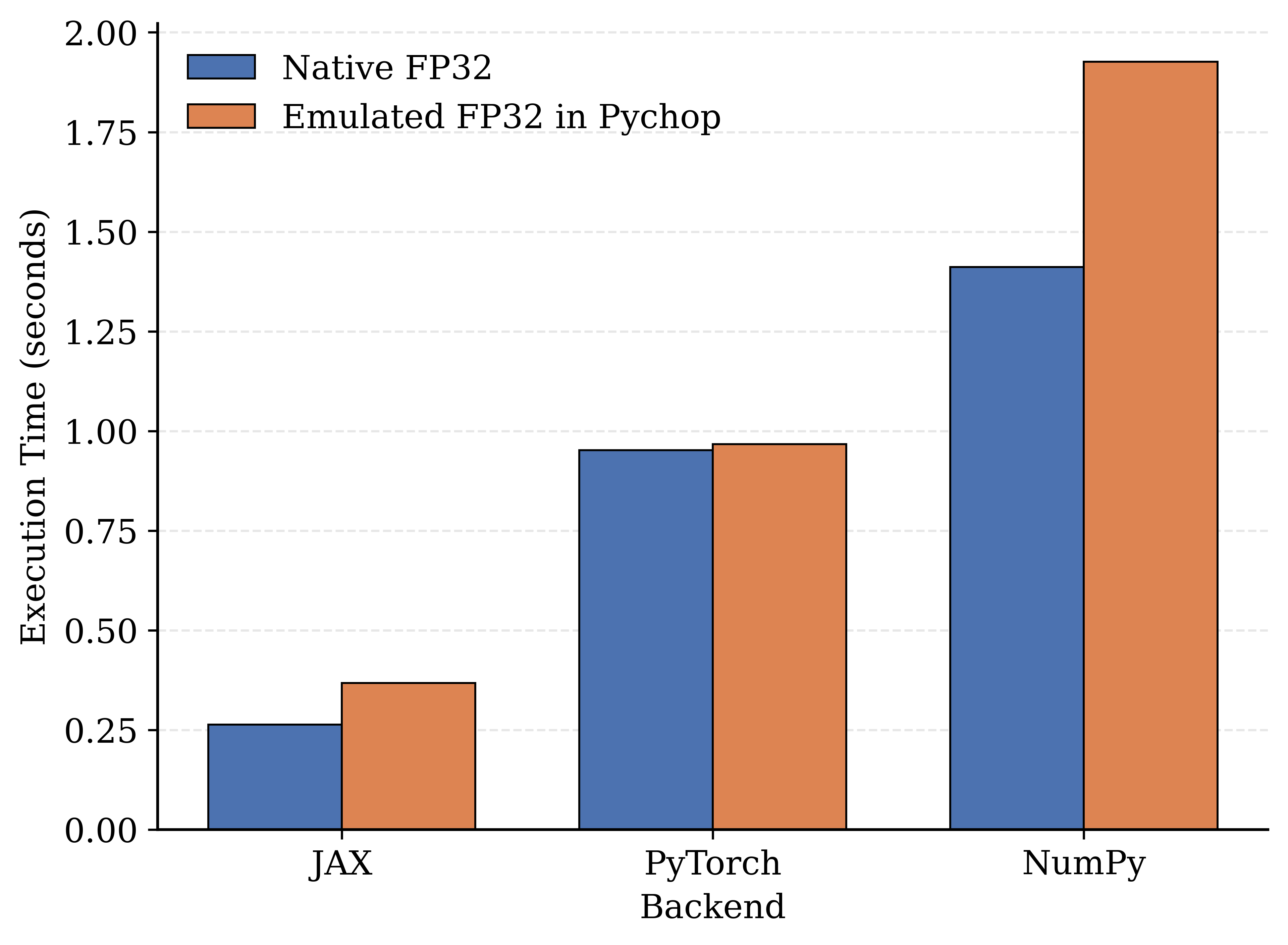}}
\caption{Runtime comparison of emulated fp32 and native fp32.}
\label{fig:emulation_comparison}
\end{figure}

The emulated variant stores data in higher precision but explicitly applies emulated fp32 chopping with round-to-nearest-even after every major floating-point operation{--}matrix-vector multiply, residual subtraction, solution update, and correction solve, while still performing the initial LU in the working precision of fp32. The emulated solution and native fp32 solution are consistent, with the average timings of four runs (discarding the first to exclude warm-up effects), and were measured on CPU for the NumPy, JAX, and PyTorch backends.  The result is as shown in \figurename~\ref{fig:emulation_comparison}.

\figurename~\ref{fig:emulation_comparison} shows that \texttt{pychop} exhibits a backend-dependent runtime penalty for emulation. The gap between emulated and native runs decreases as the data size for quantization decreases. JAX  exhibits the greatest penalty at smaller problem sizes while the relative cost decreases with scale. Except for JAX, all other backends exhibit a proportional relationship between runtime and data scale. In CPU run, NumPy performs faster than PyTorch, and the speed discrepancy narrows as the data scale increases.

\subsection{Simulations in Neural Network Quantization}
Neural network quantization (including PTQ and QAT) refers to applying reduced numerical precision (e.g., 16-bit floating-point, 8-bit integers, or even lower) in neural network training or inference instead of the standard 32-bit floating-point arithmetic typically used. In the following we explore the potential applications and advantages of utilizing \texttt{pychop} in practical scenarios.

In both quantization approaches, namely PTQ and QAT, we evaluate precisions listed in Table~\ref{table:unitroundoff1} and three customized precisions (detailed as follows) with deterministic and stochastic rounding. Stochastic rounding is rarely used in post-quantization; however, we still include it in our evaluation for completeness. For both PTQ and QAT image-classification experiments, the datasets are MNIST and FashionMNIST (standard 60,000/10,000 training/test splits), Caltech-101 (a fixed 70\% train, 15\% validation, and 15\% test split), and Oxford-IIIT Pet (37 classes, official trainval/test split).

\subsubsection{Post-quantization}

Beyond task accuracy, to evaluate the impact of quantization, we quantify the numerical errors introduced into the model parameters (weights) and intermediate feature maps (activations). Let $W$ and $A$ denote the weights and activations of the original single-precision (fp32, PyTorch default) model, respectively. Let $\hat{W}$ and $\hat{A}$ denote their quantized counterparts, respectively. We measure the distortion of all static model parameters using the Mean Squared Error (MSE) of the weights. For each parameter tensor $W^{(l)}$ in layer $l$, the per-layer MSE is defined as:
\begin{equation*}
    \text{MSE}_W^{(l)} = \frac{1}{N_l} \sum_{i=1}^{N_l} \left( W_i^{(l)} - \hat{W}_i^{(l)} \right)^2,
\end{equation*}
where $N_l$ is the number of elements in that tensor. A lower $MSE_W$ indicates a more accurate quantization format to preserve the original weight distribution. The final reported \emph{Weight MSE} is the average across all $L$ layers in the model:
\begin{equation}\label{eq:mse}
    \text{MSE} = \frac{1}{L} \sum_{l=1}^{L} \text{MSE}_W^{(l)}.
\end{equation}

While weight error captures static parameter loss, it does not account for how errors propagate through network layers. To measure dynamic signal degradation, we analyze the activations generated during inference. We employ the \emph{Signal-to-Quantization-Noise Ratio (SQNR)}, which is a robust metric for quantifying signal quality relative to quantization noise.

The SQNR (in decibels) for a given layer's activation tensor $A$ is calculated as:
\begin{equation}\label{eq:sqrn}
    \text{SQNR (dB)} = 10 \cdot \log_{10} \left( \frac{P_{signal}}{P_{noise}} \right) = 10 \cdot \log_{10} \left( \frac{\sum_{j} (A_j)^2}{\sum_{j} (A_j - \hat{A}_j)^2} \right).
\end{equation}

Here, the numerator $\sum (A_j)^2$ represents the power of the original signal, and the denominator $\sum (A_j - \hat{A}_j)^2$ represents the power of the quantization noise.

Unlike MSE, SQNR is a relative metric. Since the magnitudes of activation values vary significantly across layers of a deep neural network (e.g., initial convolution layers vs. final logits), purely absolute metrics like MSE can be dominated by layers with large numerical values. SQNR provides a scale-invariant measure of fidelity, allowing for fair comparison of quantization quality across all layers.

In our context, we compute SQNR exclusively on the final logits rather than averaging across intermediate-layer activations for several reasons. The main reason is that if we only look at the multi-layer average SQNR, it can easily overestimate or underestimate the true effect of a quantization method. Additionally, when computing SQNR for intermediate layers, the values are heavily influenced by the content of the batch (especially when class distribution is imbalanced). Focusing on the logits better captures the accumulated output effect across the model's multi-layer network. Moreover, the logits SQNR is calculated by accumulated error across the entire test set, making the result more stable by mitigating the effect of randomness. Therefore, it is more interpretable than intermediate-layer activations. Similar evaluations can also be found in \cite{10403070, 10074837}.

\paragraph{Image Classification}
This study simulates the image classification task using a pre-trained ResNet50 architecture \citep{7780459}, fine-tuned on datasets such as Caltech101 and OxfordIIITPet. For PTQ, \texttt{pychop} converts the model weights and biases, while activation SQNR is measured from the resulting quantized model outputs. In this architecture, standard layer-wise error metrics, such as MSE of intermediate activations, are often unreliable in quantized inference pipelines because of Batch Normalization (BN) folding. This causes scaling differences that can inflate error metrics and misrepresent actual information loss. In the quantized model, BN parameters are merged into the previous convolutional weights, which changes the scale of intermediate feature maps compared to the baseline. Additionally, the data augmentation techniques \texttt{RandomCrop}, \texttt{RandomHorizontalFlip}, \texttt{RandAugment} \citep{NEURIPS2020_d85b63ef}, and \texttt{Cutout} \citep{abs-1708-04552} (\texttt{n\_holes}=1, \texttt{length}=32) are employed to enhance model generalization by introducing variability in the training samples, simulating real-world image distortions.  Further, incorporating mixup data augmentation (\texttt{alpha}=1.0) \citep{ZhangCDL18} and label smoothing (0.1) \citep{10.5555/3454287.3454709} into the Cross-entropy loss function further regularizes the model, encouraging robustness against noisy labels and overfitting.

The model is trained with a batch size of 64 and a learning rate of 0.001, paired with the AdamW optimizer (weight decay = $1 \times 10^{-4}$) to support stable convergence. Training runs for 30 epochs, which is enough to adapt the pre-trained weights from ImageNet and helps prevent overfitting. A cosine annealing learning rate scheduler \citep{LoshchilovH17} gradually lowers the learning rate to improve optimization. We use mixed-precision training with PyTorch’s Automatic Mixed Precision (AMP) and GradScaler to speed up computation and reduce memory use without losing accuracy. This configuration effectively simulates the image classification task by balancing feature extraction from pre-trained weights with task-specific adaptation, achieving high test accuracies over 90\% before PTQ. We then evaluate inference after post-training quantization to e4m3, e5m2, bf16, half, tf32, as well as three custom precisions with (5,5), (5,7), and (8,4) for exponent and significand bits. The results are depicted in Tables~\ref{tab:class_result_pq} and~\ref{tab:class_result_sr}, and the classification visualizations for precisions e4m3, e5m2, bf16, half, and tf32 on Caltech101 are illustrated in \figurename~\ref{fig:imclass_all_pqt}.

\begin{figure*}[htbp]
\centering

\subfigure[e4m3 precision.]{
\includegraphics[width=0.9\linewidth]{Caltech101_q43_1_visualizations.jpg}
}

\subfigure[e5m2 precision.]{
\includegraphics[width=0.9\linewidth]{Caltech101_q52_1_visualizations.jpg}
}

\subfigure[fp16 precision.]{
\includegraphics[width=0.9\linewidth]{Caltech101_half_1_visualizations.jpg}
}

\subfigure[bf16 precision.]{
\includegraphics[width=0.9\linewidth]{Caltech101_bfloat16_1_visualizations.jpg}
}

\subfigure[tf32 precision.]{
\includegraphics[width=0.9\linewidth]{Caltech101_tf32_1_visualizations.jpg}
}

\caption{Impact of post-quantization on image classification performance on Caltech101 under different numerical precisions.}
\label{fig:imclass_all_pqt}

\end{figure*}

In the analysis of accuracy across datasets, lower-precision float types like e4m3 and e5m2 generally underperform in complex datasets under directed rounding, and e5m2 also degrades under round-toward-zero on the more complex datasets. Under directed rounding, the degradation is more severe, with accuracies as low as 1.07\% (e5m2, Caltech101, Round down) and 0.92\% (e4m3, Caltech101, Round down). In contrast, the precision Custom 1 (5,5), Custom 2 (5,7){--}especially under round-to-nearest, round-toward-zero, and stochastic rounding{--}consistently deliver high accuracies (e.g., 99.58--99.66\% on MNIST, 91.63--92.71\% on Caltech101), rivaling standard high-precision formats like half, bf16, tf32, and fp32, which stabilize at approximately 90\%--99.65\% across all datasets and rounding methods. Custom(8, 4) remains accurate under round-to-nearest, round-toward-zero, and stochastic rounding, but its performance degrades significantly under directed rounding, especially on FashionMNIST, Caltech101, and OxfordIIITPet. Notably, ``Round to nearest'' proves most reliable for maintaining accuracy across float types in PTQ, yielding high average accuracies among deterministic rounding modes (e.g., 91.64\% average accuracy for FashionMNIST), while round toward zero mode occasionally boosts custom types. Thus, custom reduced-precision formats with 5--8 exponent bits and 4--7 significand bits can provide qualified accuracy (above 90\%) under sign-symmetric or stochastic rounding for most tasks, offering an efficient trade-off between precision and performance.  In this context, sign-symmetric rounding (Round to nearest, Round toward zero, Stochastic rounding) generally performs better than directed rounding (Round up/down), particularly in low precisions.

\begin{figure}
    \centering
    \subfigure[MSE by rounding modes and precision formats]{\includegraphics[width=0.7\linewidth]{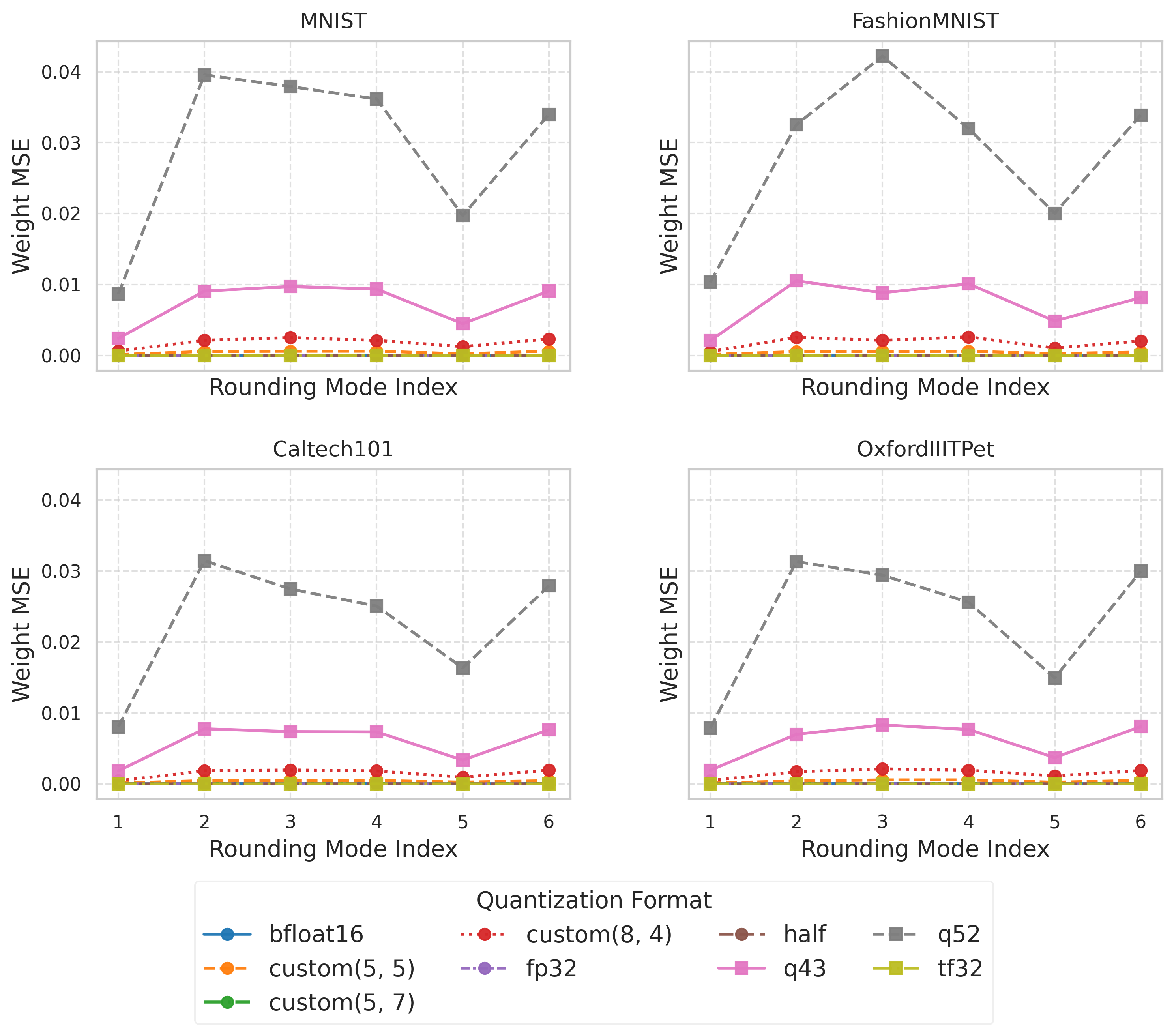}}
    \vspace{10pt}
    \subfigure[SQNR effects on accuracy.]{\includegraphics[width=0.7\linewidth]{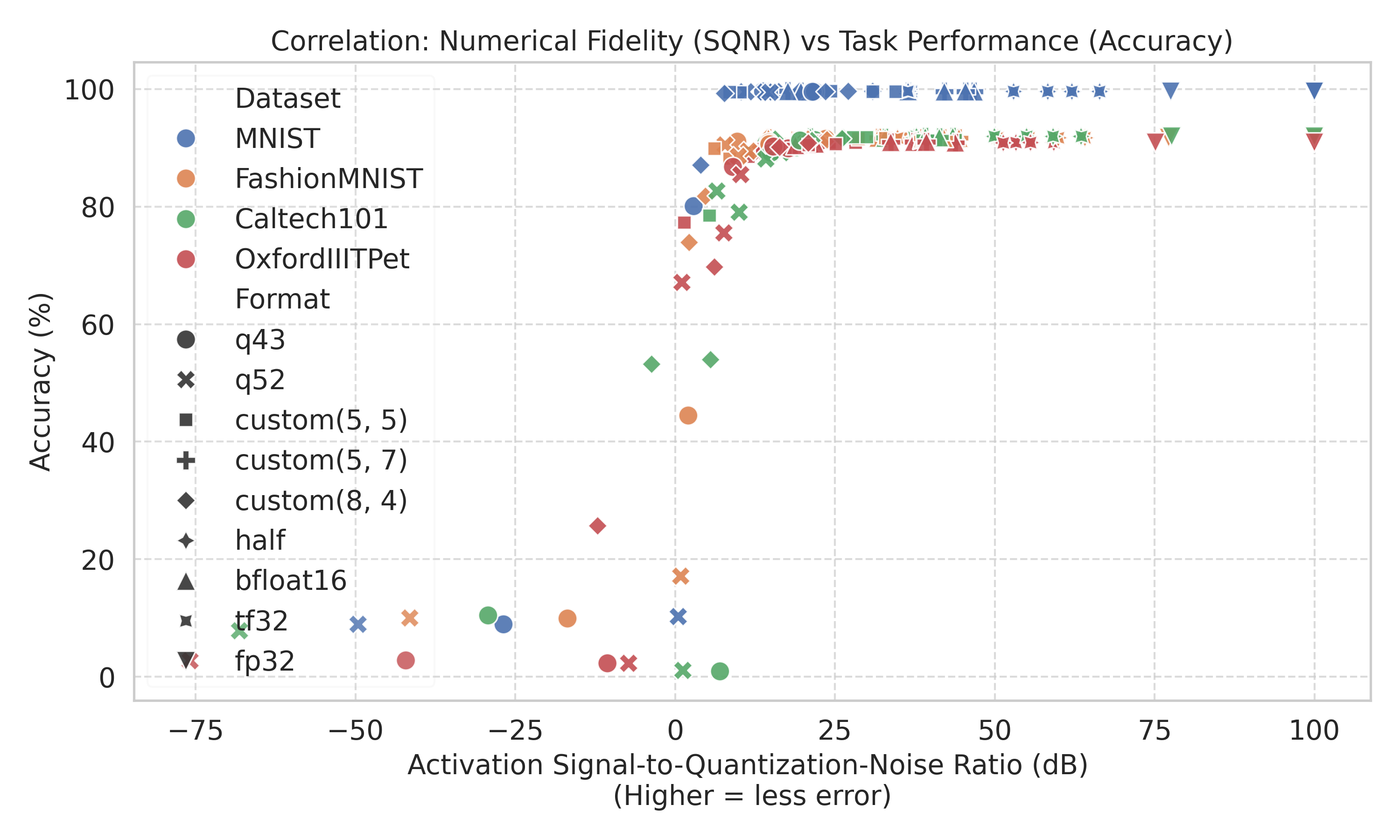}}
    \caption{Rounding impact on weight MSE and SQNR.}
    \label{fig:rounding_impact_weight_mse_sqrn}
\end{figure}

For the classification task here, SQNR is used to measure the fidelity of the final logits; we use the original FP32 model and the quantized model to perform full forward propagation on the entire test set and collect all logits for SQNR evaluation. In all datasets, round-to-nearest deterministic rounding gives the highest average SQNR among individual rounding modes, and Round up and Round down severely reduce SQNR. This trend is consistent with what we observe in accuracy: the robust modes are Round to nearest, Round toward zero, and proportional stochastic rounding, whereas directed deterministic rounding performs much worse on average. For medium- to high-precision floating-point formats (bf16, fp16, tf32, and fp32), the choice of rounding strategy has little to no impact on the final performance. In contrast, under low-precision formats such as e4m3 and e5m2, rounding to nearest in the deterministic setting is generally competitive with stochastic rounding rather than consistently better; proportional stochastic rounding often improves e4m3, while round-to-nearest remains stronger for e5m2 on complex datasets. Moreover, Round up and Round down modes exhibit severe performance degradation at these low precisions. Stochastic rounding with a uniform scheme also appears less stable on more complex datasets.  Together with \figurename~\ref{fig:rounding_impact_weight_mse_sqrn}, which depicts the Weight MSE in \eqref{eq:mse} and the SQNR effect on accuracy, we can observe that the rounding errors measured by SQNR and MSE directly affect accuracy, and that MSE is the primary factor. Additionally, once SQNR moves out of the negative or near-zero regime and MSE is controlled, accuracy increases significantly; beyond moderate SQNR levels, further increases in SQNR yield diminishing marginal gains in accuracy. This illustrates that excessively high precision contributes less to improving classification accuracy.

\begin{table}[ht]
\centering
\setlength\tabcolsep{1.2pt}
\small
\caption{Accuracy of post-quantization and activation SQNR in dB (higher is better) across datasets under different floating-point formats and deterministic rounding modes. Average row shows mean over all listed formats per rounding mode. Bold indicates the best accuracy per format across deterministic rounding modes.}
\label{tab:class_result_pq}
\begin{tabular}{ll *{4}{c@{\hspace{6pt}}c}}
\toprule
Dataset & Format & \multicolumn{2}{c}{Round to nearest} & \multicolumn{2}{c}{Round up} & \multicolumn{2}{c}{Round down} & \multicolumn{2}{c}{Round toward zero} \\
\cmidrule(lr){3-4} \cmidrule(lr){5-6} \cmidrule(lr){7-8} \cmidrule(lr){9-10}
 & & Acc. (\%) & SQNR (dB) & Acc. (\%) & SQNR (dB) & Acc. (\%) & SQNR (dB) & Acc. (\%) & SQNR (dB) \\
\midrule
\multirow{9}{*}{MNIST}
 & e4m3 & 99.56 & 21.25 & 8.92 & -26.84 & 80.08 & 2.90 & \textbf{99.59} & 13.75 \\
 & e5m2 & \textbf{99.57} & 15.45 & 8.92 & -49.59 & 10.24 & 0.49 & 99.54 & 11.09 \\
 & Custom(5,5) & 99.58 & 36.81 & 99.47 & 10.19 & 99.53 & 8.48 & \textbf{99.66} & 24.39 \\
 & Custom(5,7) & 99.58 & 46.77 & 99.58 & 19.68 & \textbf{99.65} & 17.65 & 99.62 & 36.45 \\
 & Custom(8,4) & 99.61 & 30.90 & 87.05 & 4.02 & 99.25 & 7.73 & \textbf{99.63} & 19.46 \\
 & half & 99.59 & 66.41 & 99.59 & 36.62 & \textbf{99.62} & 36.40 & 99.59 & 52.95 \\
 & bfloat16 & 99.58 & 46.77 & 99.58 & 19.68 & \textbf{99.65} & 17.65 & 99.62 & 36.45 \\
 & tf32 & 99.59 & 66.40 & 99.59 & 36.62 & \textbf{99.62} & 36.40 & 99.59 & 52.95 \\
 & fp32 & 99.58 & 100.00 & 99.58 & 100.00 & 99.58 & 100.00 & 99.58 & 100.00 \\
\cmidrule(lr){2-10}
 & Average & 99.58 & 47.20 & 78.03 & 16.71 & 88.58 & 29.08 & \textbf{99.60} & 38.83 \\
\midrule
\multirow{9}{*}{FashionMNIST}
 & e4m3 & \textbf{91.53} & 23.55 & 9.91 & -16.83 & 44.45 & 2.06 & 91.06 & 9.73 \\
 & e5m2 & 90.39 & 14.09 & 10.00 & -41.51 & 17.12 & 0.87 & \textbf{90.50} & 7.81 \\
 & Custom(5,5) & 91.63 & 35.33 & 88.20 & 8.49 & 89.92 & 6.12 & \textbf{91.73} & 19.27 \\
 & Custom(5,7) & 91.69 & 44.40 & 91.30 & 18.57 & 91.66 & 14.59 & \textbf{91.71} & 31.76 \\
 & Custom(8,4) & 91.67 & 26.18 & 73.90 & 2.20 & 81.76 & 4.75 & \textbf{91.71} & 14.45 \\
 & half & \textbf{91.71} & 64.11 & 91.60 & 34.62 & 91.67 & 34.80 & 91.69 & 50.46 \\
 & bfloat16 & 91.69 & 44.40 & 91.30 & 18.57 & 91.66 & 14.59 & \textbf{91.71} & 31.76 \\
 & tf32 & \textbf{91.71} & 64.10 & 91.60 & 34.62 & 91.67 & 34.80 & 91.69 & 50.46 \\
 & fp32 & 91.71 & 100.00 & 91.71 & 100.00 & 91.71 & 100.00 & 91.71 & 100.00 \\
\cmidrule(lr){2-10}
 & Average & \textbf{91.64} & 46.02 & 69.50 & 17.64 & 76.18 & 24.62 & 91.61 & 35.52 \\
\midrule
\multirow{9}{*}{Caltech101}
 & e4m3 & \textbf{91.33} & 21.85 & 10.44 & -29.25 & 0.92 & 7.02 & 90.71 & 14.25 \\
 & e5m2 & \textbf{89.18} & 16.62 & 7.83 & -68.15 & 1.07 & 1.21 & 82.66 & 6.53 \\
 & Custom(5,5) & \textbf{91.71} & 32.76 & 78.51 & 5.34 & 89.18 & 11.41 & 91.63 & 20.24 \\
 & Custom(5,7) & \textbf{91.94} & 43.43 & 90.10 & 17.92 & 91.79 & 21.15 & 91.71 & 32.85 \\
 & Custom(8,4) & \textbf{91.63} & 26.36 & 53.19 & -3.67 & 53.95 & 5.55 & 91.48 & 15.65 \\
 & half & 91.94 & 63.50 & 91.79 & 37.38 & 91.94 & 37.75 & 91.94 & 49.95 \\
 & bfloat16 & \textbf{91.94} & 43.43 & 90.10 & 17.92 & 91.86 & 21.15 & 91.71 & 32.85 \\
 & tf32 & 91.94 & 63.57 & 91.79 & 37.38 & 91.94 & 37.75 & 91.94 & 49.94 \\
 & fp32 & 91.94 & 100.00 & 91.94 & 100.00 & 91.94 & 100.00 & 91.94 & 100.00 \\
\cmidrule(lr){2-10}
 & Average & \textbf{91.73} & 45.72 & 67.30 & 12.76 & 67.62 & 26.33 & 90.64 & 35.25 \\
\midrule
\multirow{9}{*}{OxfordIIITPet}
 & e4m3 & \textbf{89.92} & 17.73 & 2.78 & -42.12 & 2.29 & -10.59 & 89.51 & 13.66 \\
 & e5m2 & \textbf{89.29} & 13.32 & 2.73 & -75.87 & 2.29 & -7.27 & 75.52 & 7.62 \\
 & Custom(5,5) & \textbf{90.92} & 28.21 & 77.32 & 1.42 & 88.53 & 11.47 & 90.90 & 24.12 \\
 & Custom(5,7) & 90.90 & 43.82 & 90.46 & 18.82 & 90.71 & 21.90 & \textbf{91.01} & 37.33 \\
 & Custom(8,4) & \textbf{90.84} & 22.88 & 25.67 & -12.11 & 69.72 & 6.16 & 90.73 & 19.54 \\
 & half & 90.90 & 59.24 & 90.87 & 38.27 & \textbf{90.92} & 38.22 & 90.90 & 53.32 \\
 & bfloat16 & 90.90 & 43.82 & 90.46 & 18.82 & 90.71 & 21.90 & \textbf{91.01} & 37.33 \\
 & tf32 & 90.90 & 59.25 & 90.87 & 38.27 & \textbf{90.92} & 38.21 & 90.90 & 53.31 \\
 & fp32 & 90.90 & 100.00 & 90.90 & 100.00 & 90.90 & 100.00 & 90.90 & 100.00 \\
\cmidrule(lr){2-10}
 & Average & \textbf{90.83} & 43.14 & 62.45 & 9.94 & 68.00 & 24.00 & 89.49 & 38.91 \\
\bottomrule
\end{tabular}
\end{table}

\begin{table}[ht]
\centering
\setlength\tabcolsep{3.2pt}
\small
\caption{Accuracy of post-quantization and activation SQNR in dB  (higher is better) under stochastic rounding modes. Average row shows mean over all listed formats per rounding mode. Bold indicates the best accuracy per format across stochastic rounding modes.}
\label{tab:class_result_sr}
\begin{tabular}{ll *{2}{c@{\hspace{16pt}}c}}
\toprule
Dataset & Format 
& \multicolumn{2}{c}{Stochastic (prop.)} 
& \multicolumn{2}{c}{Stochastic (uniform)} \\
\cmidrule(lr){3-4} \cmidrule(lr){5-6}
 & & Acc. (\%) & SQNR (dB) & Acc. (\%) & SQNR (dB) \\
\midrule
\multirow{9}{*}{MNIST}
 & e4m3 & \textbf{99.71} & 18.05 & 99.60 & 17.17 \\
 & e5m2 & \textbf{99.63} & 15.70 & 99.49 & 15.41 \\
 & Custom(5,5) & \textbf{99.65} & 33.73 & 99.61 & 29.73 \\
 & Custom(5,7) & \textbf{99.64} & 42.31 & 99.62 & 40.20 \\
 & Custom(8,4) & \textbf{99.70} & 21.97 & 99.63 & 23.20 \\
 & half & \textbf{99.63} & 58.23 & \textbf{99.63} & 58.57 \\
 & bfloat16 & \textbf{99.64} & 42.31 & 99.62 & 40.20 \\
 & tf32 & \textbf{99.63} & 58.23 & \textbf{99.63} & 58.57 \\
 & fp32 & \textbf{99.63} & 100.00 & \textbf{99.63} & 100.00 \\
\cmidrule(lr){2-6}
 & Average & \textbf{99.65} & 43.39 & 99.61 & 42.56 \\
\midrule
\multirow{9}{*}{FashionMNIST}
 & e4m3 & \textbf{91.71} & 19.37 & 91.32 & 17.29 \\
 & e5m2 & \textbf{91.12} & 12.65 & 89.10 & 12.78 \\
 & Custom(5,5) & \textbf{91.89} & 33.53 & 91.87 & 30.90 \\
 & Custom(5,7) & \textbf{91.91} & 33.31 & 91.84 & 39.51 \\
 & Custom(8,4) & 91.76 & 26.60 & \textbf{91.77} & 24.10 \\
 & half & \textbf{91.84} & 60.51 & \textbf{91.84} & 54.86 \\
 & bfloat16 & \textbf{91.91} & 33.31 & 91.84 & 39.51 \\
 & tf32 & \textbf{91.84} & 60.52 & \textbf{91.84} & 54.86 \\
 & fp32 & \textbf{91.84} & 100.00 & \textbf{91.84} & 100.00 \\
\cmidrule(lr){2-6}
 & Average & \textbf{91.76} & 42.20 & 91.47 & 41.53 \\
\midrule
\multirow{9}{*}{Caltech101}
 & e4m3 & \textbf{91.56} & 16.05 & 90.71 & 16.84 \\
 & e5m2 & \textbf{87.80} & 13.79 & 82.96 & 12.34 \\
 & Custom(5,5) & 92.40 & 28.10 & \textbf{92.48} & 27.97 \\
 & Custom(5,7) & 92.48 & 39.53 & \textbf{92.71} & 40.43 \\
 & Custom(8,4) & \textbf{91.94} & 25.22 & \textbf{91.94} & 21.29 \\
 & half & \textbf{92.71} & 58.99 & 92.63 & 54.94 \\
 & bfloat16 & 92.48 & 39.53 & \textbf{92.71} & 40.44 \\
 & tf32 & \textbf{92.71} & 58.98 & 92.63 & 54.96 \\
 & fp32 & \textbf{92.63} & 100.00 & \textbf{92.63} & 100.00 \\
\cmidrule(lr){2-6}
 & Average & \textbf{91.86} & 42.24 & 91.27 & 41.02 \\
\midrule
\multirow{9}{*}{OxfordIIITPet}
 & e4m3 & \textbf{90.11} & 16.73 & 87.05 & 11.65 \\
 & e5m2 & \textbf{82.77} & 6.28 & 71.35 & 7.37 \\
 & Custom(5,5) & \textbf{91.17} & 26.39 & 90.84 & 23.74 \\
 & Custom(5,7) & \textbf{91.06} & 37.55 & 91.03 & 35.92 \\
 & Custom(8,4) & \textbf{90.79} & 20.68 & 90.32 & 17.90 \\
 & half & 91.01 & 55.24 & \textbf{91.09} & 52.48 \\
 & bfloat16 & \textbf{91.06} & 37.55 & 91.03 & 35.92 \\
 & tf32 & 91.01 & 55.25 & \textbf{91.06} & 52.48 \\
 & fp32 & \textbf{91.03} & 100.00 & \textbf{91.03} & 100.00 \\
\cmidrule(lr){2-6}
 & Average & \textbf{90.00} & 39.52 & 88.31 & 37.49 \\
\bottomrule
\end{tabular}
\end{table}

\paragraph{Object Detection}
Similar to our previous task, we employed a PTQ approach to evaluate object detection performance on the COCO val2017 dataset, using various reduced-precision floating-point formats. Leveraging the \texttt{pychop} library, we applied uniform reduced-precision conversion to all neural network parameters through the \texttt{Chop} class. This method enabled us to simulate a range of floating-point formats, incorporating six rounding modes: round to nearest, round up, round down, round toward zero, and stochastic rounding (prop. and uniform). To assess object detection accuracy, we used the mAP@0.5:0.95 metric, which averages the Average Precision (AP) across Intersection over Union (IoU) thresholds from 0.5 to 0.95. This metric evaluates both detection accuracy and localization precision by measuring how well predicted bounding boxes align with ground truth boxes at varying overlap levels, with results averaged across all classes and thresholds.

\begin{figure}[ht]
\includegraphics[width=14.3cm]{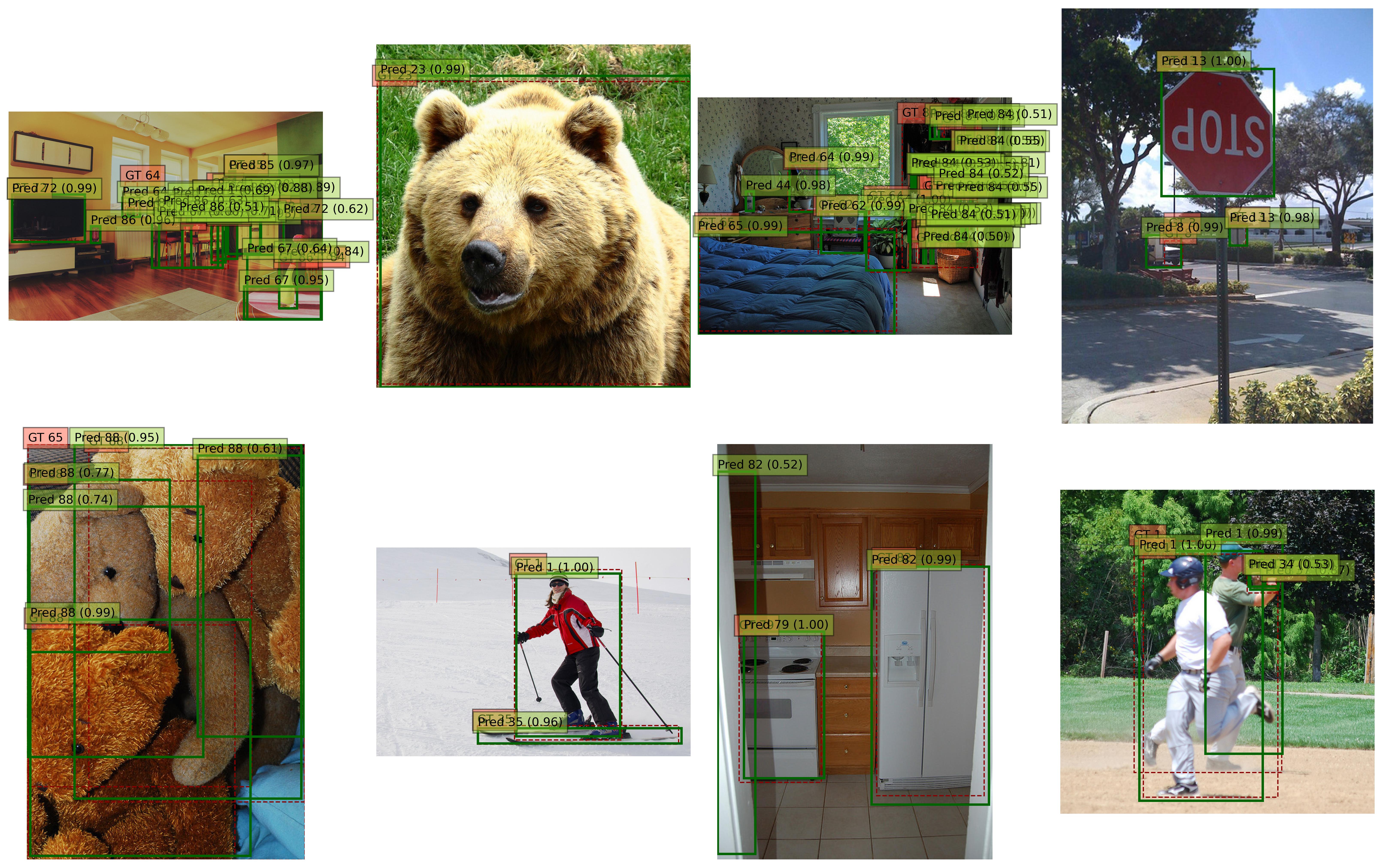}
\caption{Object detection using bf16 precision (Red: Ground Truth, Green: Predictions).}\label{fig:obj1}
\end{figure}

\begin{figure}[ht]\includegraphics[width=14.3cm]{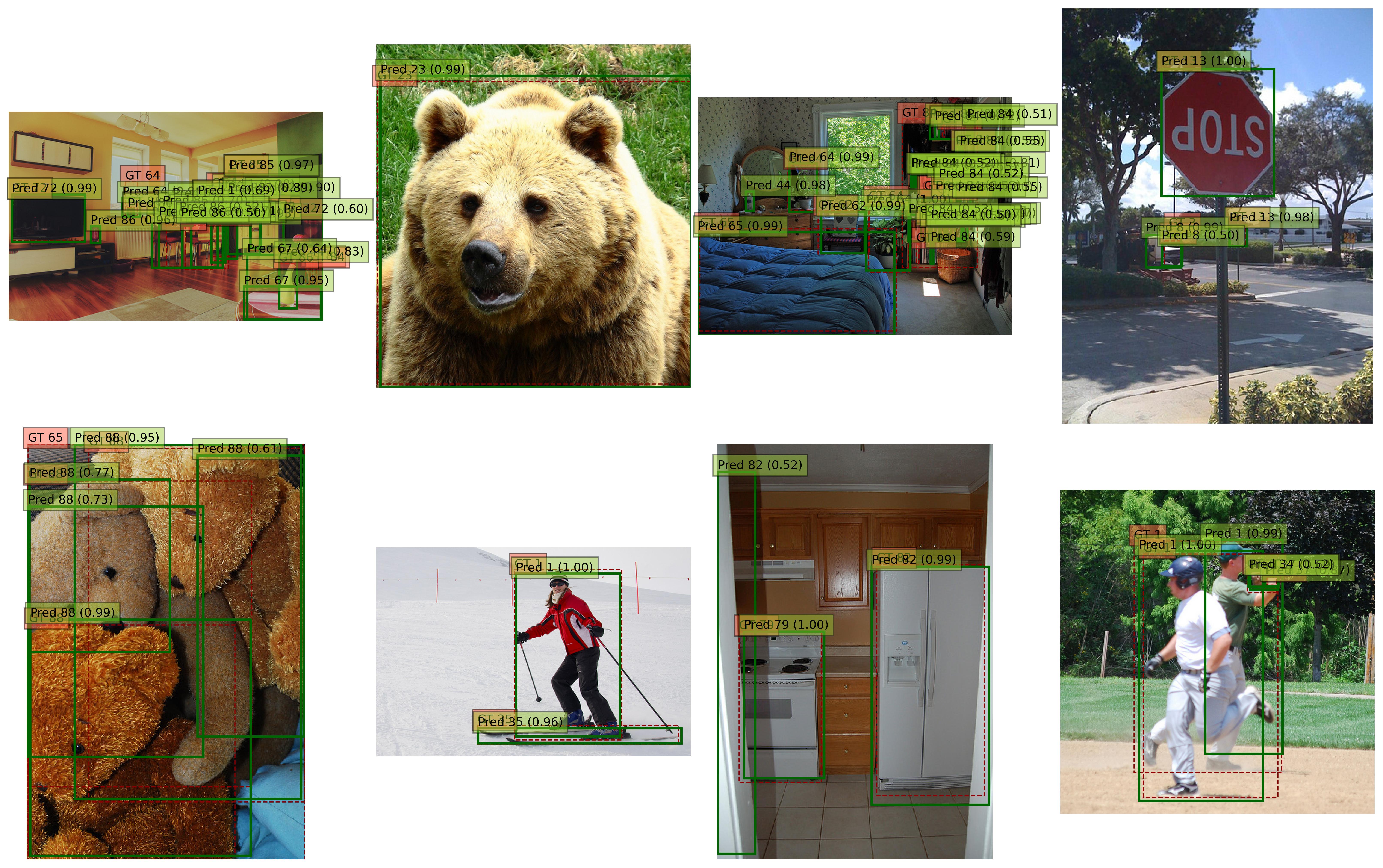}
\caption{Object detection tf32 precision (Red: Ground Truth, Green: Predictions).}\label{fig:obj2}
\end{figure}

\begin{figure}[ht]
\includegraphics[width=14.3cm]{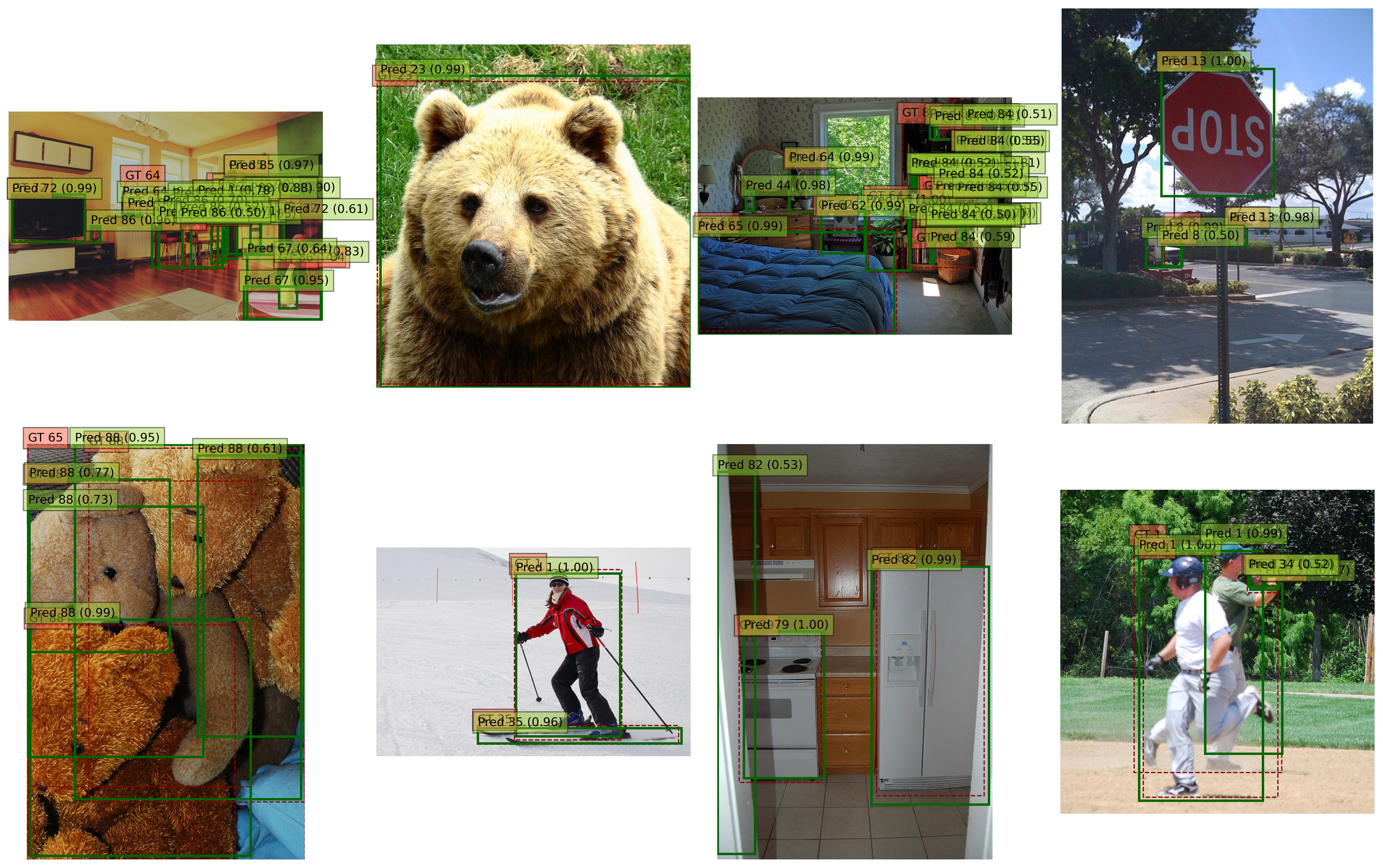}
\caption{Object detection using fp32 precision (Red: Ground Truth, Green: Predictions).}\label{fig:obj3}
\end{figure}

For this object detection task, which involves predicting object locations as bounding boxes defined by (x\_min, y\_min, width, height) alongside class labels and scores, we utilized Faster R-CNN \citep{ren2015faster} with a ResNet-50 Feature Pyramid Network (FPN) backbone \citep{lin2017feature}. The model leverages pre-trained weights from the COCO dataset, accessible through PyTorch’s \texttt{FasterRCNN\_ResNet50\_FPN\_Weights.DEFAULT}. This architecture integrates a ResNet-50 backbone for feature extraction, an FPN for multi-scale feature processing, a Region Proposal Network (RPN) for generating object proposals, and a detection head for bounding box regression and classification across 80 COCO categories plus a background class. The pre-trained weights stem from optimization on the COCO train2017 dataset ($\sim$118,000 images) over 12 epochs with a batch size of 2, employing a composite loss: the RPN combines binary cross-entropy loss for objectness classification with Smooth L1 loss for proposal regression, while the detection head uses cross-entropy loss for classification and Smooth L1 loss for box refinement, guided by a step-wise learning rate schedule (e.g., 0.02 initial rate, decayed at epochs 8 and 11). In our experiments, we applied these weights directly for inference on a subset of 100 images from COCO val2017, as specified by \texttt{max\_images=100}, bypassing additional training. The \texttt{post\_quantization} function from \texttt{pychop} converted the model weights and biases into the specified precisions, allowing us to evaluate the impact of reduced-precision emulation on mAP@0.5:0.95 across the defined rounding strategies. The results are as depicted in \tablename~\ref{tab:obj_quant_combined} and visualized in \figurename~\ref{fig:obj1}, \figurename~\ref{fig:obj2}, and \figurename~\ref{fig:obj3}, respectively.
For the object detection evaluation with Faster R-CNN, the number of RPN proposal regions is variable and the ROI head shapes change dynamically, making direct comparison of the final outputs using SQNR infeasible. Therefore, we use the feature maps of the backbone FPN as representative activation values to calculate SQNR as in \eqref{eq:sqrn}.  Specifically, we register forward hooks on the FPN and extract the P2 feature map (the highest-resolution level) from a batch of 8 images. Similar to the classification task, MSE is evaluated with \eqref{eq:mse}.

In general, higher mAP is typically associated with lower MSE and higher SQNR, showing a negative correlation with MSE and a positive correlation with SQNR. Among deterministic rounding modes, Round to nearest achieves the highest mAP across almost all precision formats, particularly in low-precision ones such as e4m3 (mAP=0.449) and e5m2 (0.433). As shown in Tables~\ref{tab:obj_quant_combined} and~\ref{tab:obj_quant_stochastic}, both exponent bit and significand bit contribute to improved SQNR scores. When the significand bit reaches a certain level (e.g., 5), then the exponent bit plays a critical role since the increase of the exponent bit can prevent information loss for small values, e.g., activation vanishing. As shown in Tables~\ref{tab:obj_quant_combined} and~\ref{tab:obj_quant_stochastic}, Custom(5,5), Custom(5,7), bf16, half, and tf32 achieve similar scores under round-to-nearest and stochastic rounding, whereas Custom(8,4), e4m3, and e5m2 are more sensitive to directed rounding and show substantial degradation under Round up/down. In particular, the custom precision Custom(5,5) achieves the highest score even with a relatively lower SQNR value compared to fp32, which acts as an outlier here.

Symmetric or less biased rounding methods, such as Round to nearest, Round toward zero, and Stochastic (prop.), significantly outperform deterministic directed rounding methods (such as Round up/down) under low precision. This is consistent with the situation we previously discussed for the classification problem. In neural network deployment, the directed rounding (`Round up'' or `Round down'') can cause activation attenuation or distribution shift, leading to degraded or unstable inference and amplified rounding errors, particularly in the context of PTQ and low-precision formats (e.g., e4m3 and e5m2).  Due to directional bias, the asymmetric quantization of positive and negative weights results in linearly accumulating errors, which destroys the balance of the model (for example, activation values may be overly amplified or attenuated).

Round toward zero performs better than Round up/down under low precision but is still inferior to Round to nearest; for example, under e4m3 it achieves an mAP of 0.282 (far higher than up/down's 0 score), and under Custom(5,5) it reaches 0.438 (close to nearest's 0.457). Round toward zero also yields higher SQNR (e.g., Custom(5,5) 18.76 dB), because it treats positive and negative numbers symmetrically, causing quantization errors to converge toward zero.

For stochastic rounding, Stochastic (prop.) generally outperforms Stochastic (uniform), especially in low-precision scenarios (e.g., e5m2: mAP=0.410 vs. uniform 0.350; e4m3: 0.413 vs. uniform 0.000), while uniform is only marginally better for tf32 and half (0.451 vs. 0.450), and fp32 ties at 0.450. Similarly, it is observed that Stochastic (prop.) performs far better than deterministic up/down.  Consistent with our earlier discussion, we find that in medium-to-high precision formats (bf16, tf32, half, fp32), rounding has a much smaller effect on mAP than in low-precision formats, with performance mostly around 0.450; however, directed rounding still lowers SQNR for bf16 to 17.40--18.68 dB, while tf32, half, and fp32 retain MSE close to 0 and SQNR generally above 30 dB.

\begin{table}[htp]
\centering
\setlength\tabcolsep{1.2pt}
\small
\caption{Object detection performance (mAP@0.5:0.95, higher is better), weight MSE (lower is better) and feature SQNR in dB (higher is better) under different floating-point formats and deterministic rounding methods. Bold indicates the best mAP per format across rounding modes. Three significant digits are preserved. We mark the values smaller than $1\times 10^{-8}$ as zeros.}
\label{tab:obj_quant_combined}
\begin{tabular}{l *{4}{c@{\hspace{2pt}}c@{\hspace{2pt}}c}}
\toprule
& \multicolumn{3}{c}{Round to nearest} 
& \multicolumn{3}{c}{Round up} 
& \multicolumn{3}{c}{Round down} 
& \multicolumn{3}{c}{Round toward zero} \\
\cmidrule(lr){2-4} \cmidrule(lr){5-7} \cmidrule(lr){8-10} \cmidrule(lr){11-13}
Format & mAP & MSE & SQNR (dB) 
& mAP & MSE & SQNR (dB) 
& mAP & MSE & SQNR (dB) 
& mAP & MSE & SQNR (dB) \\
\midrule
e4m3          & \textbf{0.449} & \(1.18\times10^{-6}\) & 20.97 
             & 0.000 & \(4.68\times10^{-6}\) & \(-\infty\) 
             & 0.000 & \(4.55\times10^{-6}\) & -99.59 
             & 0.282 & \(4.20\times10^{-6}\) & 7.42 \\

e5m2          & \textbf{0.433} & \(3.65\times10^{-6}\) & 15.72 
             & 0.000 & \(1.44\times10^{-5}\) & -56.95 
             & 0.000 & \(1.67\times10^{-5}\) & 2.26 
             & 0.023 & \(1.52\times10^{-5}\) & 4.57 \\

Custom (5,5) & \textbf{0.457} & \(7.00\times10^{-8}\)  & 31.10 
             & 0.203 & \(2.40\times10^{-7}\)  & 5.00 
             & 0.366 & \(2.40\times10^{-7}\)  & 8.55 
             & 0.438 & \(2.40\times10^{-7}\)  & 18.76 \\

Custom (5,7) & \textbf{0.451} & \(0\)                 & 46.63 
             & 0.441 & \(2.00\times10^{-8}\)    & 17.40 
             & 0.444 & \(2.00\times10^{-8}\)    & 18.68 
             & 0.443 & \(2.00\times10^{-8}\)    & 30.99 \\

Custom (8,4) & \textbf{0.444} & \(2.70\times10^{-7}\)  & 26.27 
             & 0.025 & \(9.0\times10^{-7}\)  & -2.04 
             & 0.189 & \(9.7\times10^{-7}\)  & 5.47 
             & 0.409 & \(9.2\times10^{-7}\)  & 13.21 \\

bf16         & \textbf{0.451} & \(0\)                 & 46.63 
             & 0.441 & \(2.00\times10^{-8}\)    & 17.40 
             & 0.444 & \(2.00\times10^{-8}\)    & 18.68 
             & 0.443 & \(2.00\times10^{-8}\)    & 30.99 \\

tf32         & 0.450 & \(0\)                 & 62.72
             & \textbf{0.451} & \(0\)                 & 35.83 
             & 0.448 & \(0\)                 & 35.92 
             & 0.450 & \(0\)                 & 48.31 \\

half         & 0.450 & \(0\)                 & 62.72 
             & \textbf{0.451} & \(0\)                 & 35.83 
             & 0.448 & \(0\)                 & 35.92 
             & 0.450 & \(0\)                 & 48.31 \\

fp32         & \textbf{0.450} & \(0\)                 & 131.37 
             & \textbf{0.450} & \(0\)          & 131.38 
             & \textbf{0.450} & \(0\)          & 130.17 
             & \textbf{0.450} & \(0\)          & 130.16 \\

\bottomrule
\end{tabular}
\end{table}

\begin{table}[htp]
\centering
\setlength\tabcolsep{1.2pt}
\small
\caption{Object detection performance (mAP@0.5:0.95, higher is better), weight MSE (lower is better) and feature SQNR in dB (higher is better) under stochastic rounding methods.  Bold indicates the best mAP per format across stochastic rounding modes. Three significant digits are preserved. We mark the values smaller than $1\times 10^{-8}$ as zeros.}
\label{tab:obj_quant_stochastic}
\begin{tabular}{l *{2}{c@{\hspace{16pt}}c@{\hspace{16pt}}c}}
\toprule
& \multicolumn{3}{c}{Stochastic (prop.)} 
& \multicolumn{3}{c}{Stochastic (uniform)} \\
\cmidrule(lr){2-4} \cmidrule(lr){5-7}
Format & mAP & MSE & SQNR (dB) 
& mAP & MSE & SQNR (dB) \\
\midrule
e4m3          & \textbf{0.413} & \(2.69\times10^{-6}\) & 16.73 
             & 0.000 & \(4.31\times10^{-6}\) & -120.34 \\

e5m2          & \textbf{0.410} & \(6.76\times10^{-6}\) & 12.24 
             & 0.350 & \(1.43\times10^{-5}\) & 9.64 \\

Custom (5,5) & \textbf{0.452} & \(1.40\times10^{-7}\)  & 30.24 
             & 0.444 & \(2.10\times10^{-7}\)  & 27.81 \\

Custom (5,7) & \textbf{0.453} & \(1.00\times10^{-8}\)    & 42.63 
             & 0.448 & \(1.00\times10^{-8}\)    & 40.08 \\

Custom (8,4) & \textbf{0.448} & \(4.50\times10^{-7}\)  & 25.65 
             & 0.442 & \(1.08\times10^{-6}\)  & 22.34 \\

bf16         & \textbf{0.453} & \(1.00\times10^{-8}\)    & 42.63 
             & 0.448 & \(1.00\times10^{-8}\)    & 40.08 \\

tf32         & 0.450 & \(0\)                 & 59.50 
             & \textbf{0.451} & \(0\)                 & 56.76 \\

half         & 0.450 & \(0\)                 & 59.50 
             & \textbf{0.451} & \(0\)                 & 56.76 \\

fp32         & \textbf{0.450} & \(0\)                 & 131.38 
             & \textbf{0.450} & \(0\)                 & 130.17 \\

\midrule
Average      & 0.442 & \(1.12\times10^{-6}\) & 46.72 
             & 0.387 & \(2.21\times10^{-6}\) & 29.26 \\

\bottomrule
\end{tabular}
\end{table}

% directed rounding errors will be absorbed for quantization aware training - increase the accuracy of uniform stochastic rounding

\subsubsection{Quantization-aware Training}
We perform quantization-aware training of ResNet-18 with \texttt{pychop} to simulate its effect under different reduced-precision floating-point formats. The backbone is initialized with ImageNet-pretrained weights from \texttt{torchvision\footnote{https://github.com/pytorch/vision}}, with the first convolutional layer adapted to each dataset's input channel count (1 for grayscale datasets, 3 otherwise) by averaging pretrained weights across channels when necessary.

\begin{table}[ht]
\centering\setlength\tabcolsep{1.2pt}\small
\caption{Accuracy of Quantization-aware Training Across Datasets and Float Types with Different Rounding Modes}\label{tab:class_result_qat} 
\begin{tabular}{llcccccc}
\toprule
Dataset & Format &{Round to nearest} & {Round up} & {Round down} & {Round toward zero} & {Stochastic (prop.)} & {Stochastic (uniform)}  \\
\midrule
\multirow{10}{*}{MNIST} 
 & e4m3         & 99.28\% & 99.26\% & 99.28\% & 99.28\% & \textbf{99.29\%} & 99.21\% \\
 & e5m2         & \textbf{99.33\%} & 99.18\% & 99.31\% & 99.29\% & 99.27\% & 99.28\% \\
 & Custom 1 (5, 5) & 99.27\% & 99.21\% & 99.30\% & 99.32\% & 99.28\% & \textbf{99.40\%} \\
 & Custom 2 (5, 7) & \textbf{99.32\%} & 99.22\% & 99.27\% & 99.31\% & 99.24\% & 99.26\% \\
 & Custom 3 (8, 4) & \textbf{99.44\%} & 99.31\% & 99.26\% & 99.36\% & 99.30\% & 99.28\% \\
 & half         & 99.30\% & \textbf{99.31\%} & 99.26\% & 99.30\% & 99.29\% & 99.25\% \\
 & bf16         & 99.24\% & 99.28\% & \textbf{99.29\%} & 99.26\% & 99.26\% & 99.21\% \\
 & tf32         & 99.25\% & \textbf{99.40\%} & 99.27\% & 99.23\% & 99.28\% & 99.33\% \\
 & fp32         & 99.32\% & 99.29\% & 99.28\% & 99.28\% & \textbf{99.33\%} & 99.28\% \\
 & Average      & \textbf{99.31\%} & 99.27\% & 99.28\% & 99.29\% & 99.28\% & 99.28\% \\
\midrule
\multirow{10}{*}{FashionMNIST} 
 & e4m3         & \textbf{91.11\%} & 90.97\% & 91.02\% & 91.09\% & 91.02\% & 90.99\% \\
 & e5m2         & 91.07\% & 91.19\% & 91.13\% & \textbf{91.24\%} & 91.18\% & 90.92\% \\
 & Custom 1 (5, 5) & 91.18\% & 91.15\% & 91.16\% & 91.04\% & 91.12\% & \textbf{91.27\%} \\
 & Custom 2 (5, 7) & 91.16\% & 91.15\% & 91.01\% & \textbf{91.39\%} & 90.84\% & 91.30\% \\
 & Custom 3 (8, 4) & \textbf{91.19\%} & 90.89\% & 91.06\% & \textbf{91.19\%} & 90.99\% & \textbf{91.19\%} \\
 & half         & 91.11\% & \textbf{91.41\%} & 91.05\% & 91.21\% & 91.15\% & 91.19\% \\
 & bf16         & 91.09\% & 91.14\% & 91.20\% & 90.99\% & \textbf{91.31\%} & 91.17\% \\
 & tf32         & 91.10\% & 91.09\% & 91.08\% & 90.95\% & \textbf{91.44\%} & 91.09\% \\
 & fp32         & 91.13\% & \textbf{91.16\%} & 90.98\% & 90.98\% & 90.92\% & 90.86\% \\
 & Average      & 91.13\% & \textbf{91.13\%} & 91.08\% & 91.12\% & 91.11\% & 91.11\% \\
\midrule
\multirow{10}{*}{Caltech101} 
 & e4m3         & 89.26\% & 89.18\% & 90.18\% & \textbf{90.25\%} & 88.95\% & 88.95\% \\
 & e5m2         & \textbf{89.72\%} & 87.11\% & 86.80\% & 89.64\% & 88.87\% & 88.33\% \\
 & Custom 1 (5, 5) & \textbf{89.64\%} & 89.49\% & 89.33\% & 89.10\% & \textbf{89.64\%} & 89.10\% \\
 & Custom 2 (5, 7) & 89.49\% & 88.87\% & \textbf{90.18\%} & 89.56\% & 89.41\% & 89.49\% \\
 & Custom 3 (8, 4) & \textbf{90.48\%} & 89.03\% & 89.87\% & 89.03\% & 89.03\% & 89.64\% \\
 & half         & 89.64\% & 89.95\% & 90.10\% & 88.95\% & \textbf{90.56\%} & 89.56\% \\
 & bf16         & 89.33\% & \textbf{90.33\%} & 89.26\% & 89.64\% & 89.49\% & 90.02\% \\
 & tf32         & 88.87\% & \textbf{91.17\%} & 89.26\% & 89.95\% & 90.02\% & 88.87\% \\
 & fp32         & 89.18\% & \textbf{89.33\%} & 89.18\% & 89.26\% & 89.26\% & 89.18\% \\
 & Average      & \textbf{89.51\%} & 89.38\% & 89.35\% & 89.49\% & 89.47\% & 89.24\% \\
\midrule
\multirow{10}{*}{OxfordIIITPet} 
 & e4m3         & \textbf{87.14\%} & 86.35\% & 86.62\% & 86.10\% & \textbf{87.14\%} & 85.96\% \\
 & e5m2         & 86.54\% & 81.58\% & 83.86\% & 86.48\% & \textbf{86.67\%} & 85.94\% \\
 & Custom 1 (5, 5) & 86.67\% & \textbf{86.81\%} & 86.64\% & 86.35\% & \textbf{86.81\%} & 86.32\% \\
 & Custom 2 (5, 7) & 86.94\% & 86.48\% & \textbf{87.16\%} & 86.45\% & 86.67\% & 86.24\% \\
 & Custom 3 (8, 4) & 86.70\% & 86.07\% & 86.64\% & \textbf{86.73\%} & 86.40\% & 86.54\% \\
 & half         & \textbf{87.05\%} & 86.94\% & 86.78\% & 86.78\% & 86.21\% & 86.54\% \\
 & bf16         & 86.59\% & \textbf{86.94\%} & 86.81\% & 86.84\% & 86.10\% & 86.54\% \\
 & tf32         & \textbf{87.14\%} & 86.67\% & 86.94\% & 86.51\% & 86.48\% & 86.59\% \\
 & fp32         & \textbf{86.75\%} & \textbf{86.75\%} & \textbf{86.75\%} & \textbf{86.75\%} & \textbf{86.75\%} & \textbf{86.75\%} \\
 & Average      & \textbf{86.84\%} & 86.07\% & 86.47\% & 86.55\% & 86.58\% & 86.38\% \\
\bottomrule
\end{tabular}
\end{table}

For quantization-aware training with \texttt{pychop}, we replace all Conv2d and Linear layers with custom modules that insert fake quantization. Weights are quantized in all layers, while activations are quantized in all convolutional layers except the initial one (to avoid quantizing raw inputs) and are omitted before the final classifier.

Using the shared dataset setup described above, we evaluate precisions e4m3, e5m2, bf16, half, tf32, as well as three custom precisions with (5,5), (5,7), and (8,4) exponent and significand bits, with subnormal numbers disabled. Six rounding modes provided by \texttt{pychop} are tested for each. Training uses AdamW with a learning rate of $10^{-3}$, weight decay of $10^{-4}$, and cosine annealing over 10 epochs with a batch size of 64. For Caltech-101, the train and validation subsets described above are combined for QAT training.  Data augmentation (random resized crop, horizontal flip, normalization, and cutout for grayscale datasets) is applied. The reported accuracy corresponds to the best checkpoint obtained during the training run. For qualitative analysis, we generate visualizations of the first 20 test samples with prediction probabilities and save quantized models for each configuration. The prediction plots for the first 20 examples in the test set are shown in \figurename~\ref{fig:imclass_all_qat}.

In terms of the results depicted in \tablename~\ref{tab:class_result_qat}, for simple tasks (MNIST and FashionMNIST), the differences among rounding methods are negligible. In contrast, for complex tasks (Caltech101 and OxfordIIITPet), the rounding effect is significantly amplified. Round-to-nearest provides a competitive overall average performance, particularly on MNIST, Caltech101, and Oxford-IIIT Pet; however, it does not eliminate the competitiveness of stochastic rounding in 8-bit precision, with Stochastic (prop.) achieving the best or near-best accuracy in several aggressive formats such as e4m3 and e5m2. On FashionMNIST, the differences among the leading rounding modes are negligible. The Stochastic (uniform) rounding no longer exhibits severe degradation and is generally close to other rounding modes, although its average accuracy remains slightly below the strongest deterministic modes on the Caltech101 and OxfordIIITPet datasets.
For the higher-precision formats (bf16, tf32, half, and fp32), rounding has a relatively small and less systematic effect, allowing most rounding methods to be chosen freely.  Stochastic (proportional) enables the model to achieve near full-precision performance even under extremely low-precision formats such as e4m3 and e5m2. We observe that QAT exhibits substantially greater tolerance to different rounding methods than PTQ. In this study, Round up and Round down perform much better than they did under the PTQ strategy. Nevertheless, to fully unlock the potential of low-precision quantization, Round to nearest is the most robust empirical choice in our setting, while unbiased stochastic rounding (Stochastic (prop.)) remains a competitive alternative.  The potential advantages of Stochastic rounding (prop.) come largely from its unbiased feature, which prevents tiny gradient and activation values vanishing during the training and avoids overfitting. This can be explained by the fact that QAT applies fake quantization in the forward pass while the STE propagates gradients during the backward pass, allowing model parameters to adapt to quantization noise throughout training and compensate for quantization errors. In addition, stochastic rounding can act as implicit regularization, as stochastic rounding transforms quantization error into high-frequency random noise rather than systematic bias.  Similar to the regularization effect introduced by dropout or label smoothing, this noise brought by the stochastic rounding helps the model escape shallow local optima and improves generalization capability.  Similar to the regularization effect introduced by dropout or label smoothing, this noise brought by the stochastic rounding helps the model escape shallow local optima and improves generalization capability. PTQ applies quantization only once at inference time and lacks this adaptability of model parameters to the introduced noise; a similar outcome and explanation can be found in \cite{jacob2018quantization}.

Our empirical results for stochastic rounding echo those of existing studies on training neural networks (e.g., \cite{10.5555/3045118.3045303, liu-etal-2025-training}); in neural network training (particularly in reduced-precision or quantized training), stochastic rounding is generally beneficial because it effectively eliminates the systematic bias introduced by deterministic rounding \citep{10.1098/rsos.211631}, prevents small updates from being consistently discarded, and allows rounding errors to tend toward positive and negative cancellation (i.e., they cancel each other out on average). \cite{10.5555/3045118.3045303} demonstrates the use of stochastic rounding in training deep networks using only 16-bit wide fixed-point number representation and shows little to no degradation in classification accuracy.

These results suggest that the benefit of stochastic rounding is context-dependent; distance-proportional stochastic rounding can be competitive in low precision{--}especially when quantization noise can be absorbed during training, and it is much less harmful than directed rounding when directed bias dominates. In contrast, uniform stochastic rounding may inject excessive noise in PTQ, particularly for low-precision object detection.

\begin{figure*}[htbp]
\centering

\subfigure[e4m3 precision.]{
\includegraphics[width=0.9\linewidth]{quant_Caltech101_q43_1_visualizations.jpg}
}

\subfigure[e5m2 precision.]{
\includegraphics[width=0.9\linewidth]{quant_Caltech101_q52_1_visualizations.jpg}
}

\subfigure[bf16 precision.]{
\includegraphics[width=0.9\linewidth]{quant_Caltech101_bfloat16_1_visualizations.jpg}
}

\subfigure[fp16 precision.]{
\includegraphics[width=0.9\linewidth]{quant_Caltech101_half_1_visualizations.jpg}
}

\subfigure[tf32 precision.]{
\includegraphics[width=0.9\linewidth]{quant_Caltech101_tf32_1_visualizations.jpg}
}

\caption{Impact of quantization-aware training on image classification performance on Caltech101 under different numerical precisions.}
\label{fig:imclass_all_qat}
\end{figure*}

\section{Software Development}
\label{sec:software_version_scope}

\texttt{pychop} has continued to evolve as an open-source software package. The experimental evaluation and API examples in this paper should be interpreted as using \texttt{pychop} version 0.6.0. This release extends the software coverage to include OCP MX formats, block floating-point formats, Flexpoint-style block formats, and additional backend support such as TensorFlow\footnote{https://www.tensorflow.org/}. These capabilities improve support for emerging low-precision formats and deployment workflows; however, OCP MX, BFP, Flexpoint-style formats, and TensorFlow-specific experiments are not included in the numerical results reported in Section~\ref{sec:exps}. A systematic empirical evaluation of these formats across all supported backends is left for future work.

\section{Conclusion}\label{sec:conclusion}
In this work, we present \texttt{pychop} for efficient reduced-precision emulation for numerical methods and deep learning applications. By seamlessly integrating automatic differentiation support into PyTorch and JAX, \texttt{pychop} enhances accessibility and usability for computational scientists working with different software frameworks. Its flexibility, customizable design, and comprehensive rounding support enable advanced mixed-precision numerical algorithms and deep learning applications.

Empirical results across various rounding modes and precisions in Python and MATLAB for reduced-precision floating-point emulation demonstrated that \texttt{pychop} achieves a competitive speedup over MATLAB \texttt{chop}, with improvements of multiple orders of magnitude when deployed on a GPU. In addition, we simulated post-quantization and quantization-aware training effects for image classification on the MNIST, FashionMNIST, Caltech101, and OxfordIIITPet datasets, as well as object detection on the COCO dataset, to further explore its usage in deep learning deployments in reduced-precision arithmetic. These experiments offer valuable insights into the optimal bit widths required for exponents and significands to achieve high-quality inference, highlighting performance trade-offs to guide the selection of efficient floating-point representations tailored to specific use cases. The empirical results confirm that \texttt{pychop} delivers practical performance across modern frameworks, with GPU acceleration providing the largest gains for large-scale computations typical in deep learning workloads.

\bibliographystyle{ACM-Reference-Format}
\bibliography{refs}

\appendix

\section{Appendix}

\subsection{Floating-point Precision Emulation}\label{app:fpt}
\begin{center}
\begin{minipage}[t]{0.45\textwidth}
\begin{lstlisting}[title={Pass precision format (fp16) directly}]
from pychop import FaultChop
import numpy as np

X = np.random.randn(5000, 5000)
ch = FaultChop('h', rmode=1, flip=True, subnormal=True) # Standard IEEE 754 half precision
# other parameters are left as default.
Xq = ch(X) # Rounding values
\end{lstlisting}
\end{minipage}
\hspace{0.05\textwidth}
\begin{minipage}[t]{0.45\textwidth}
\begin{lstlisting}[title={Customized precision}]
import numpy as np
from pychop import FaultChop, Chop
from pychop import Customs

X = np.random.randn(5000, 5000)
ch = FaultChop(customs=Customs(exp_bits=5, sig_bits=10), rmode=1) # half precision (5 exponent bits, 10+(1) significand bits, (1) is implicit bits)
Xq = ch(X)

ch = Chop(exp_bits=5, sig_bits=10, rmode=1, subnormal=True)
Xq = ch(X)
\end{lstlisting}
\end{minipage}
\end{center}

\subsection{Fixed-point Precision Emulation}\label{app:fp}
\begin{lstlisting}[title={Fix-point representation}]
from pychop import Chopf
import numpy as np

X = np.random.randn(5000, 5000)
ch = Chopf(ibits=4, fbits=4, rmode=1)
ch(X)
\end{lstlisting}

\subsection{Integer Quantization}\label{app:iqt}
\begin{lstlisting}
import numpy as np
from pychop import Chopi

X_np = np.random.randn(5000, 5000)
ch = Chopi(bits=8, symmetric=False, per_channel=False, axis=0)
X_q = ch.quantize(X_np) # to integers
X_inv = ch.dequantize(X_q) # back to floating-point numbers
\end{lstlisting}

\subsection{Mathematical Function}\label{app:math_func}
The usage of common functions is illustrated as an example below:
\begin{lstlisting}
import numpy as np
import pychop.math_func as mf
from pychop import Chop

chopper = Chop(exp_bits=5, sig_bits=10, rmode=3)
x = np.array([0.0, 1.5708])
result = mf.sin(x, chopper)
print(result)
\end{lstlisting}

\subsection{Backend Specification}\label{app:backend}

In the example below, there are three distinct inputs: a NumPy array, a PyTorch tensor, and a JAX array. By specifying backend, we can configure \texttt{pychop} to use 5 exponent bits and 10 significand bits with round-to-nearest mode to process these inputs accordingly. For instance, selecting the ``torch'' backend allows \texttt{pychop} to handle \texttt{X\_th}, the PyTorch tensor. GPU deployment can be enabled by transferring the PyTorch tensor or JAX array to a GPU device, such as with ``\texttt{X\_th.to('cuda')}'' when CUDA is available.

\begin{lstlisting}
import numpy as np
import torch
import jax.numpy as jnp
import pychop
from pychop import Chop

X_np = np.random.randn(5000, 5000) # Numpy array
X_th = torch.Tensor(X_np) # torch array
X_jx = jnp.asarray(X_np)

pychop.backend('numpy') # Use NumPy backend explicitly.
ch = Chop(exp_bits=5, sig_bits=10, rmode=1)
emulated= ch(X_np)

pychop.backend('torch') # Use PyTorch backend.
ch = Chop(exp_bits=5, sig_bits=10, rmode=1)
emulated= ch(X_th)

pychop.backend('jax') # Use JAX backend.
ch = Chop(exp_bits=5, sig_bits=10, rmode=1)
emulated= ch(X_jx)
\end{lstlisting}

\subsection{Quantization-aware Training}\label{app:qat}

In the following, we demonstrate how to use the derived layer modules in \texttt{pychop.layers} (following \tablename~\ref{tab:layers}) to build quantization-aware training for convolutional neural networks.
\begin{center}
\begin{minipage}[t]{0.45\textwidth} % , style=PythonCode
\begin{lstlisting}[title={Use built-in precision}]
import torch.nn as nn
import torch.nn.functional as F

class CNN(nn.Module):
    def __init__(self):
        super().__init__()
        self.conv1 = nn.Conv2d(1, 16, 3, 1, 1)
        self.pool = nn.MaxPool2d(2, 2)
        self.conv2 = nn.Conv2d(16, 32, 3, 1, 1)
        self.fc1 = nn.Linear(32 * 7 * 7, 128)
        self.fc2 = nn.Linear(128, 10)

    def forward(self, x):
        x = F.relu(self.conv1(x))
        x = self.pool(x)
        x = F.relu(self.conv2(x))
        x = self.pool(x)
        x = x.view(-1, 32 * 7 * 7)
        x = F.relu(self.fc1(x))
        x = self.fc2(x)
        return x
\end{lstlisting}
\end{minipage}
\hspace{0.05\textwidth}
\begin{minipage}[t]{0.45\textwidth}% , style=PythonCode
\begin{lstlisting}[title={Specify quantizer.}]
import torch.nn as nn
import torch.nn.functional as F
from pychop.layers import QuantizedConv2d, QuantizedMaxPool2d, QuantizedLinear

class QuantizedCNN(nn.Module):
    def __init__(self, chop1, chop2):
        super().__init__()
        self.conv1 = QuantizedConv2d(1, 16, 3, chop=chop1)
        self.pool = QuantizedMaxPool2d(2, chop=chop1)
        self.conv2 = QuantizedConv2d(16, 32, 3, chop=chop1)
        self.fc1 = QuantizedLinear(32 * 5 * 5, 128, chop=chop2)
        self.fc2 = QuantizedLinear(128, 10, chop=chop2)

    def forward(self, x):
        x = F.relu(self.conv1(x))
        x = self.pool(x)
        x = F.relu(self.conv2(x))
        x = self.pool(x)
        x = x.view(x.size(0), -1)
        x = F.relu(self.fc1(x))
        x = self.fc2(x)
        return x
\end{lstlisting}
\end{minipage}
\end{center}

\subsection{Reduced-precision Optimizers}
For reduced-precision optimizers, similarly, one can define the derived class of optimizers, as in the example below.
\begin{lstlisting}[title={Customized reduced-precision optimization for neural network deployment.}]
import torch.nn as nn
from pychop import ChopSTE
from pychop.layers import QuantizedLinear, QuantizedReLU
from pychop.optimizers import (
    QuantizedSGD,
    QuantizedAdam,
    QuantizedRMSprop,
    QuantizedAdagrad
)

# Simple quantized model for demonstration (MLP)
class QuantizedMLP(nn.Module):
    """Simple 3-layer MLP with quantized layers for QAT demonstration."""
    def __init__(self, chop=None):
        super().__init__()
        self.fc1 = QuantizedLinear(784, 256, chop=chop)
        self.relu1 = QuantizedReLU(chop=chop)
        self.fc2 = QuantizedLinear(256, 128, chop=chop)
        self.relu2 = QuantizedReLU(chop=chop)
        self.fc3 = QuantizedLinear(128, 10, chop=chop)

    def forward(self, x):
        x = x.view(x.size(0), -1)
        x = self.relu1(self.fc1(x))
        x = self.relu2(self.fc2(x))
        x = self.fc3(x)
        return x

if __name__ == "__main__":
    # Define reduced-precision quantizers (different rounding modes)
    chop_low = ChopSTE(exp_bits=5, sig_bits=10, rmode=1)
    chop_mid = ChopSTE(exp_bits=5, sig_bits=10, rmode=4)

    # Create model with reduced-precision QAT enabled
    model = QuantizedMLP(chop=chop_low)

    # Customized reduced-precision quantized optimizers
    optimizer_sgd = QuantizedSGD(
        model.parameters(), lr=0.01, momentum=0.9, chop=chop_low
    )

    optimizer_rmsprop = QuantizedRMSprop(
        model.parameters(), lr=0.01, chop=chop_mid
    )

    optimizer_adam = QuantizedAdam(
        model.parameters(), lr=0.001, chop=chop_low
    )

    optimizer_adagrad = QuantizedAdagrad(
        model.parameters(), lr=0.01, chop=chop_mid
    )

    print("Model and optimizers ready for reduced-precision QAT training.")
\end{lstlisting}

\subsection{Support in MATLAB}\label{app:matlab}

To trigger the Python virtual environment, one must have Python and the \texttt{pychop} library installed (e.g., via pip manager using \texttt{pip install pychop}), then simply pass the following command in your MATLAB terminal:
\begin{lstlisting}[style=MATLAB-editor]
>> pe = pyenv()  % or specify your python environment by ``pe = pyenv('Version', '/software/python/anaconda3/bin/python3')``
\end{lstlisting}

%\begin{boxedverbatim}

% \end{boxedverbatim}
\begin{framed}
\begin{verbatim}[fontsize=\footnotesize] % pe =
  PythonEnvironment with properties:
          Version: "3.10"
       Executable: "/software/python/anaconda3/bin/python3"
          Library: "/software/python/anaconda3/lib/libpython3.10.so"
             Home: "/software/python/anaconda3"
           Status: NotLoaded
    ExecutionMode: InProcess
\end{verbatim}
\end{framed}

To use \texttt{pychop} in your MATLAB environment, similarly, simply load the \texttt{pychop} module:

\begin{lstlisting}
np = py.importlib.import_module('numpy');
pc = py.importlib.import_module('pychop');

X = np.asarray(rand(100, 100));
ch = pc.Chop(pyargs('exp_bits', int32(5), 'sig_bits', int32(10), 'rmode', int32(1)));
X_q = ch(X);
\end{lstlisting}

\end{document}